\documentclass[11pt]{article}

\usepackage[margin=1in]{geometry}
\usepackage{amsmath}
\usepackage{amssymb}
\usepackage{bm}
\usepackage{newtxtext}

\usepackage{algorithm}
\usepackage{algpseudocode}
\usepackage{booktabs}
\usepackage{graphicx}
\usepackage[hidelinks]{hyperref}

\DeclareMathOperator*{\argmax}{arg\,max}

\DeclareMathOperator{\sigmoid}{sig}
\DeclareMathOperator{\logit}{logit}
\DeclareMathOperator{\softplus}{sfp}

\newcommand{\authorblock}[3]{%
  \begin{tabular}[t]{c}
    \textbf{#1} \\
    \small #2 \\
    \small \texttt{#3}
  \end{tabular}%
}

\title{\textbf{Nonlocal Transition Kernel for Efficient Learning of Restricted Boltzmann Machines}}

\author{
  \authorblock{Kaiji Sekimoto}{Graduate School of Science and Engineering, Yamagata University, Japan}{sekimoto@yz.yamagata-u.ac.jp}
  \and
  \authorblock{Muneki Yasuda}{Graduate School of Science and Engineering, Yamagata University, Japan}{muneki@yz.yamagata-u.ac.jp}
}

\date{}

\begin{document}

\maketitle

\begin{abstract}
Learning restricted Boltzmann machines (RBMs) is computationally challenging because it requires expectations whose exact evaluation is generally intractable. The expectations are typically evaluated using a sampling approximation based on blocked Gibbs sampling (BGS), which is a local Markov chain Monte Carlo transition kernel. However, the locality of BGS can lead to poor sampling quality when the RBM has high energy barriers, thereby degrading learning performance. Deep tempering (DT), which performs parallel tempering over a sequence of learnable RBMs including the training RBM, alleviates this locality issue. However, DT algorithmically requires multiple steps to move through the RBM sequence to achieve a nonlocal transition. In this paper, we propose a transition kernel defined over the RBM sequence used in DT. The proposed kernel has a round-trip structure over the sequence, enabling nonlocal moves within a single transition while leaving the RBM sequence invariant. Numerical experiments show that the proposed kernel performs nonlocal transitions more frequently and achieves higher sampling quality with fewer transitions than BGS and DT. We further verify that learning based on the proposed kernel is more stable and mitigates the training failures observed with BGS- and DT-based learning.
\end{abstract}

\section{Introduction} \label{sec:introduction}

A restricted Boltzmann machine (RBM) is a latent-variable energy-based model with a two-layer architecture and has been applied to various tasks such as pretraining deep learning models~\cite{salakhutdinov2010_preTrain}, collaborative filtering~\cite{salakhutdinov2007_collaborative,Abdollahi2016_collaborative}, hierarchical clustering~\cite{Decelle2023_hierarchical}, inferring effective many-body interactions~\cite{Decelle2024_coupling}, anomaly detection~\cite{fiore2013_anomaly,Do2018_anomaly,Sekimoto2024_ad}, analysis in many-body physics~\cite{carleo2017_phy,lu2019_phy}, and design of polymer materials~\cite{Hatakeyama2022_design}. RBM training is performed using maximum likelihood estimation and requires the evaluation of expectations over the RBM. However, the evaluation is generally difficult and requires an approximation method. Accurate approximation is essential for achieving high learning performance. Since blocked Gibbs sampling (BGS), a type of Markov chain Monte Carlo (MCMC), can be easily performed for an RBM, the expectations are typically evaluated using a sampling approximation based on BGS. 

The sampling approximation directly depends on the sampling quality, that is, how closely the sample distribution follows the target distribution. Although MCMC theoretically requires infinitely many transition steps to sample from the target distribution, this is impractical, and the transition must be terminated after a finite number of steps. Therefore, in practice, the resulting sample distribution does not exactly match the target distribution, and it is thus desirable for the finite-step sample distribution to approximate the target distribution as closely as possible. This can be achieved through two design choices: (i) an initial distribution that roughly approximates the target distribution, and (ii) a rapidly mixing transition kernel. The contrastive divergence (CD) method~\cite{Hinton2002_rbm} and the persistent CD (PCD) method~\cite{Tieleman2008_PCD}, which are well-known RBM learning methods, corresponds to the former approach. They set the initial distribution to the data distribution and the sample distribution used in the previous parameter update, respectively. The sampling quality of these learning methods also directly depends on the mixing time of BGS. BGS is a local transition kernel that explores states in the neighborhood of the current state, and transitions between modes are difficult when the target distribution is multimodal and the modes are separated by high energy barriers. As a result, the transition strongly depends on the initial state, and thus the mixing time may grow dramatically. Therefore, when an RBM is trained on data distributions with such a structure, the mixing time of BGS on the RBM gradually increases, degrading the sampling quality. Since the degradation of the sampling quality is attributed to the locality of BGS, an additional transition kernel capable of global exploration is required to achieve rapid mixing during RBM training.

A class of nonlocal transition kernels can be constructed using a sequence of tempered distributions with gradually increasing temperatures. This class includes parallel tempering~\cite{swendsen1986_ReMC,hukushima1996_ReMC}, simulated tempering~\cite{Marinari1992_ST}, and tempered transition~\cite{Neal1996_TT,Behrens2010_TT}. The temperature parameter controls the flatness of the target distribution, and higher-temperature distributions can generally be explored more rapidly, even using local transition kernels. Parallel tempering and simulated tempering combine two types of updates: local transitions within each tempered distribution and moves between distributions along the temperature ladder. By traversing the sequence of tempered distributions, these methods can facilitate nonlocal exploration. These methods leave the entire sequence, including the target distribution, invariance. Parallel tempering has been applied to the RBM training and has been shown to yield more stable training than methods based on BGS~\cite{Desjardins2010_PTTrain,Cho2010_PTTrain}. Deep tempering (DT) has also been proposed as a learning method based on parallel tempering~\cite{desjardins2014_deeptempering}. Whereas parallel tempering constructs a sequence of distributions by varying the temperature, DT represents the distributions in the sequence using learnable models and obtains the sequence by jointly training these models with the training RBM. DT has been shown to achieve learning performance comparable to that of parallel tempering using a smaller number of intermediate distributions~\cite{desjardins2014_deeptempering}. In parallel tempering and DT, transitions between distributions are restricted to adjacent distributions in the temperature or model sequence. In simulated tempering, transitions between distributions are also typically implemented only between adjacent temperatures. Consequently, moving a state from a rapidly mixing high-temperature distribution to the target distribution requires multiple transitions. It is therefore algorithmically difficult for these methods to produce a nonlocal state-space move within a single transition. In contrast, tempered transition is a Metropolis--Hastings algorithm whose proposal is generated by a round-trip transition from the target distribution to the highest-temperature distribution and back. By passing through high-temperature distributions, it can readily cross energy barriers and thus has the potential to produce a nonlocal move within a single transition. However, when such a proposal is rejected, the chain remains at its current state and fails to make even a local move. Therefore, tempered transition is designed to achieve a sufficiently high acceptance probability. Although this can be achieved using a finer temperature schedule, doing so increases the computational cost. Based on the foregoing, we aim to construct a transition kernel that preserves an invariant distribution over a sequence of distributions, as do parallel tempering, DT, and simulated tempering, while retaining the potential of tempered transition to perform nonlocal moves within a single transition.

In this paper, we propose a novel transition kernel defined on the sequence of RBMs used in DT. The proposed transition kernel has a round-trip structure analogous to that of tempered transitions and thus could perform a nonlocal state-space move within a single transition. However, unlike tempered transitions, the proposed kernel does not rely on the Metropolis--Hastings algorithm and therefore does not suffer from the drawback of a nonlocal proposal being rejected, causing the chain to remain at the current state. We further ensure that the proposed transition kernel leaves the RBM sequence invariant, as does DT. We numerically demonstrate that the proposed kernel produces nonlocal moves more frequently than BGS and DT. As a result, it maintains high sampling quality even with a relatively small number of transition steps. We show that the proposed method achieves high learning performance. In particular, the proposed method mitigates learning failures in settings where learning based on BGS and DT typically fails.

The remainder of this paper is organized as follows. In Section~\ref{sec:rbm}, we introduce the RBM and its learning procedure, and numerically investigate training failures and their underlying causes. In Sections~\ref{sec:DT} and \ref{sec:proposed}, we introduce DT and the proposed transition kernel with nonlocal moves, respectively. In Section~\ref{sec:num_exp}, we evaluate the sampling and learning performance of the proposed method through numerical experiments on some synthetic and real-world datasets. Finally, in Section~\ref{sec:conclusion}, we summarize our findings and discuss potential directions for future work.

\section{Restricted Boltzmann Machines and Their Learning} \label{sec:rbm}

\subsection{Restricted Boltzmann Machines }

\begin{figure}[t]
    \centering
    \includegraphics[width=0.45\linewidth]{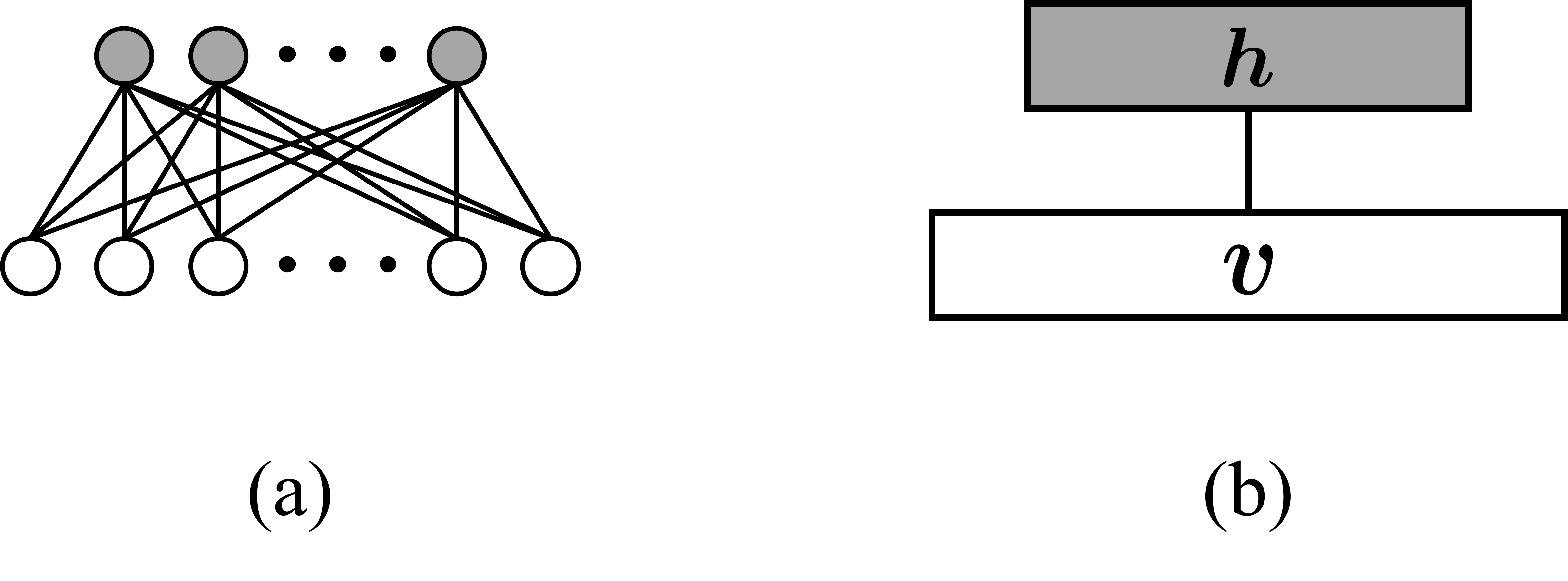}
    \caption{Schematic illustrations of an RBM: (a) a two-layer architecture with a visible layer at the bottom and a hidden layer at the top, and (b) its simplified representation.}
    \label{fig:rbm}
\end{figure}

An RBM is a Markov random field defined on a complete bipartite graph~\cite{Smolensky1986_rbm,Hinton2002_rbm}. The graph structure can be regarded as a two-layer architecture consisting of a visible layer, to which the visible variables are assigned, and a hidden layer, to which the hidden variables are assigned. Figure \ref{fig:rbm} illustrates the RBM structure. Consider an RBM with $n$ visible variables $\bm{v}\in\{0,1\}^{n}$ and $m$ hidden variables $\bm{h}\in\{0,1\}^{m}$. The distribution of the RBM is defined as
\begin{align}
P_{\theta}(\bm{v},\bm{h}) := \frac{1}{Z_{\theta}} \exp\left[-E(\bm{v},\bm{h};\theta)\right], 
\label{eq:rbm}
\end{align}
where $E$ denotes the energy function defined by
\begin{align}
E(\bm{v},\bm{h};\theta) := - \bm{b}^\top \bm{v} - \bm{c}^\top \bm{h} - \bm{v}^\top \bm{W} \bm{h},
\label{eq:energy}
\end{align}
and $Z_{\theta}$ denotes the normalizing constant (or the partition function) expressed as
\begin{align*}
Z_{\theta} := \sum_{\bm{v}} \sum_{\bm{h}}\exp\left[-E(\bm{v},\bm{h};\theta)\right],
\end{align*}
where $\sum_{\bm{v}}$ and $\sum_{\bm{h}}$ denote multiple summations over all possible realizations of the visible and hidden variables, respectively. Here, $\bm{b}\in\mathbb{R}^{n}$ and $\bm{c}\in\mathbb{R}^{m}$ denote the bias parameters for the visible and hidden variables, respectively, and $\bm{W}\in\mathbb{R}^{n\times m}$ denotes the connection parameters between the visible and hidden variables. These are learnable parameters and are collectively denoted by
\begin{align*}
\theta
:=
\begin{pmatrix}
\bm{b} \\
\bm{c} \\
\mathrm{vec}(\bm{W})
\end{pmatrix}
\in
\mathbb{R}^{n+m+n m},
\end{align*}
where $\mathrm{vec}(\bm{W})$ denotes the column vector obtained by arranging the elements of $\bm{W}$ according to a predefined order. The visible variables $\bm{v}$ correspond to a data point, and the number of the visible variables, $n$, is thus equal to the data dimension. By contrast, the hidden variables $\bm{h}$ are model-internal latent variables that do not directly correspond to a data point. The number of hidden variables, $m$, controls the representational power of the RBM, and increasing the number of hidden variables enhances this power~\cite{Le2008_rbmPower}. The conditional distributions for the visible and hidden variables are written, respectively, as
\begin{align}
P_{\theta}(\bm{v}\mid\bm{h}) &= \prod_{i=1}^{n} \mathrm{Bern}\bigl(v_{i}\mid \sigmoid(\eta_i(\bm{h};\theta))\bigr), \label{eq:rbm_v|h} \\
P_{\theta}(\bm{h}\mid\bm{v}) &= \prod_{j=1}^{m} \mathrm{Bern}\bigl(h_{j}\mid \sigmoid(\zeta_j(\bm{v};\theta))\bigr), \label{eq:rbm_h|v}
\end{align}
where $\mathrm{Bern}(x\mid p) = p^x (1-p)^{1-x}$ is a Bernoulli distribution with the probability $p\ge 0$ that the random variable $x$ takes the value one. Here, $\sigmoid(x):=1/(1+e^{-x})$ is the sigmoid function, and
\begin{align*}
\eta_i(\bm{h};\theta) &:= b_{i} + \sum_{j=1}^{m} W_{i,j} h_{j}, \\
\zeta_j(\bm{v};\theta) &:= c_{j} + \sum_{i=1}^{n} W_{i,j} v_{i}.
\end{align*}
The conditional distributions in Eqs.~\eqref{eq:rbm_v|h} and \eqref{eq:rbm_h|v} show that the variables in one layer are conditionally independent given the variables in the other layer. This conditional independence facilitates BGS between the two layers. 

Given a current joint state $(\bm{v},\bm{h})$, one BGS sweep first samples a hidden state from $P_{\theta}(\bm{h}'\mid\bm{v})$ and subsequently samples $\bm{v}'$ from $P_{\theta}(\bm{v}'\mid\bm{h}')$ given the generated hidden state. The resulting transition kernel on the joint state space is
\begin{align*}
K_{\theta}(\bm{v}',\bm{h}'\mid\bm{v},\bm{h})
&:= P_{\theta}(\bm{h}'\mid\bm{v}) P_{\theta}(\bm{v}'\mid\bm{h}').
\end{align*}
Although its right-hand side does not depend on the current hidden state $\bm{h}$, we retain $\bm{h}$ in the conditioning argument so that $K_{\theta}$ can be regard as a Markov transition kernel on the joint state space of $\bm{v}$ and $\bm{h}$. The kernel $K_\theta$ leaves the RBM joint distribution in Eq.~\eqref{eq:rbm} invariant because it satisties the balance condition expressed as 
\begin{align*}
\sum_{\bm{v}}\sum_{\bm{h}}
K_{\theta}(\bm{v}',\bm{h}'\mid\bm{v},\bm{h})
P_{\theta}(\bm{v},\bm{h})
&= P_{\theta}(\bm{v}',\bm{h}').
\end{align*}
When only the visible variables are retained as the Markov-chain state, the induced transition kernel on the visible state space is
\begin{align}
T_{\theta}^{\mathrm{BGS}}(\bm{v}'\mid\bm{v})
&:= \sum_{\bm{h}'} K_{\theta}(\bm{v}',\bm{h}'\mid\bm{v},\bm{h}) \nonumber \\
&= \sum_{\bm{h}'} P_{\theta}(\bm{v}'\mid\bm{h}') P_{\theta}(\bm{h}'\mid\bm{v}). \label{eq:bgs}
\end{align}
Here, $\bm{h}'$ serves as an auxiliary variable for the visible-state Markov chain. The kernel $T_{\theta}^{\mathrm{BGS}}$ leaves the marginal visible distribution in Eq.~\eqref{eq:rbm_v} invariant because it satisfies the detailed balance condition written by
\begin{align*}
T_{\theta}^{\mathrm{BGS}}(\bm{v}'\mid\bm{v}) P_{\theta}(\bm{v}) 
&= T_{\theta}^{\mathrm{BGS}}(\bm{v}\mid\bm{v}') P_{\theta}(\bm{v}') .
\end{align*}
As described in Appendix~\ref{app:trans_matrix}, the relaxation time, which characterizes the timescale over which the influence of the initial state decays, is determined from the second-largest eigenvalue of the transition probability matrix. For a finite Markov chain satisfying the detailed balance condition, the relaxation time appears in an upper bound on the mixing time~\cite{David2017_MCMC}. Thus, the second-largest eigenvalue is an important measure for evaluating the sampling performance of BGS in Eq.~\eqref{eq:bgs}.

\subsection{Training Restricted Boltzmann Machines}

Suppose that we obtain a dataset consisting of $N_{\mathrm{d}}$ data points corresponding to $\bm{v}$ expressed as
\begin{align*}
\mathfrak{D} := \{\mathbf{d}^{(\mu)}\in\{0,1\}^{n}\mid \mu=1,2,\ldots,N_{\mathrm{d}}\}
\end{align*}
For the dataset $\mathfrak{D}$, we consider the data distribution expressed as
\begin{align}
Q_\mathfrak{D}(\bm{v}) = \frac{1}{N_{\mathrm{d}}}\sum_{\mu=1}^{N_{\mathrm{d}}} \delta(\bm{v},\mathbf{d}^{(\mu)}),
\label{eq:data_dist}
\end{align}
where $\delta(\cdot)$ denotes the Kronecker delta function. The log-likelihood function is then defined as
\begin{align}
\ell(\theta) := \mathbb{E}_{Q_\mathfrak{D}(\bm{v})} [\ln P_{\theta}(\bm{v})] = \frac{1}{N_{\mathrm{d}}}\sum_{\mu=1}^{N_{\mathrm{d}}} \ln P_{\theta}^{(v)}(\mathbf{d}^{(\mu)}), \label{eq:log_likelihood}
\end{align}
where $ \mathbb{E}_{Q_\mathfrak{D}(\bm{v})} [\cdot]$ denotes the expectation over the data distribution, and $P_{\theta}^{(v)}(\bm{v})$ is the marginal distribution for the visible variables represented as
\begin{align}
P_{\theta}^{(v)}(\bm{v}) 
&= \frac{1}{Z_{\theta}} \exp\left[-\hat{E}(\bm{v};\theta)\right],
\label{eq:rbm_v}
\end{align}
where $\hat{E}$ denotes the energy function for the visible distribution defined by
\begin{align}
\hat{E}(\bm{v};\theta) := - \bm{b}^\top \bm{v} - \sum_{j=1}^{m}\softplus\bigl(\zeta_j(\bm{v};\theta)\bigr),
\label{eq:energy_v}
\end{align}
where $\softplus(x):=\ln(1+e^x)$ is the softplus function. RBM training is performed by maximizing the log-likelihood in Eq.~\eqref{eq:log_likelihood} with respect to the learning parameters $\theta$. The maximization is typically performed using gradient ascent, which iteratively updates the parameters according to
\begin{align*}
\theta^{\mathrm{new}} \leftarrow \theta^{\mathrm{old}} + \varepsilon \nabla_{\theta} \ell(\theta)|_{\theta = \theta^{\mathrm{old}}}, 
\end{align*}
where $\varepsilon>0$ denotes a small positive learning rate, and $\theta^{\mathrm{old}}$ and $\theta^{\mathrm{new}}$ denote the values of $\theta$ before and after the update, respectively. The parameter gradient, $\nabla_{\theta} \ell$, is written as
\begin{align}
\nabla_{\theta} \ell(\theta) = - \mathbb{E}_{Q_\mathfrak{D}(\bm{v})}[\nabla_{\theta} \hat{E}(\bm{v};\theta)]  + \mathbb{E}_{P_{\theta}(\bm{v}, \bm{h})}\bigl[\nabla_{\theta} E(\bm{v}, \bm{h};\theta)\bigr], \label{eq:grad}
\end{align}
where $\mathbb{E}_{P_{\theta}(\bm{v}, \bm{h})}[\cdots]$ denotes the expectation over the RBM distribution in Eq.~\eqref{eq:rbm} defined by
\begin{align}
\mathbb{E}_{P_{\theta}(\bm{v}, \bm{h})}[f(\bm{v}, \bm{h})] := \sum_{\bm{v}} \sum_{\bm{h}} f(\bm{v}, \bm{h}) P_{\theta}(\bm{v}, \bm{h}). \label{eq:expectation}
\end{align}
In particular, the gradients with respect to the parameters, $b_i$ , $c_j$, and $W_{i,j}$, are represented as
\begin{align}
\frac{\partial\,\ell(\theta)}{\partial\, b_i} &= \mathbb{E}_{Q_\mathfrak{D}(\bm{v})}[v_i] - \mathbb{E}_{P_{\theta}(\bm{v}, \bm{h})}[v_i], \label{eq:grad_b} \\
\frac{\partial\,\ell(\theta)}{\partial\, c_j} &= \mathbb{E}_{Q_\mathfrak{D}(\bm{v})}\left[ \sigmoid\bigl(\zeta_j(\bm{v};\theta)\bigr)\right] - \mathbb{E}_{P_{\theta}(\bm{v}, \bm{h})}[h_j], \label{eq:grad_c} \\
\frac{\partial\,\ell(\theta)}{\partial\, W_{i,j}} &= \mathbb{E}_{Q_\mathfrak{D}(\bm{v})}\left[ v_i \sigmoid\bigl(\zeta_j(\bm{v};\theta)\bigr)\right] - \mathbb{E}_{P_{\theta}(\bm{v}, \bm{h})}[v_i h_j], \label{eq:grad_W}
\end{align}
respectively. Because the expectation in Eq.~\eqref{eq:expectation} involves the multiple summations, $\sum_{\bm{v}}$ and $\sum_{\bm{h}}$, which cause the combinatorial explosion, maximizing Eq.~\eqref{eq:log_likelihood} using gradient ascent is computationally difficult. Therefore, the expectation must be evaluated using an approximation method in practice. To properly maximize the log-likelihood using gradient ascent, it is important to accurately estimate the expectation appearing in the parameter gradient.

The expectation is typically evaluated using a sampling approximation based on BGS. Let us consider a sample distribution based on BGS, expressed as
\begin{align*}
P_{\mathrm{BGS}}^s(\bm{v},\bm{h}) := P_\theta(\bm{h}\mid\bm{v}) \sum_{\bm{v}'} T_{\theta,s}^{\mathrm{BGS}}(\bm{v}\mid\bm{v}') P_0(\bm{v}'),
\end{align*}
where $P_0$ and $T_{\theta,s}^{\mathrm{BGS}}$ denote the initial distribution and the $s$-step transition kernel of BGS in Eq.~\eqref{eq:bgs}, respectively. We then generate $N_{\mathrm{s}}$ sample points from the sample distribution, and the expectation in Eq.~\eqref{eq:expectation} is approximated by
\begin{align*}
\mathbb{E}_{P_{\theta}(\bm{v}, \bm{h})}[f(\bm{v}, \bm{h})] \approx \frac{1}{N_{\mathrm{s}}}\sum_{\nu=1}^{N_{\mathrm{s}}} f(\mathbf{v}^{(\nu)}, \mathbf{h}^{(\nu)}),
\end{align*}
where $\{(\mathbf{v}^{(\nu)}, \mathbf{h}^{(\nu)})\mid \nu=1,2,\ldots,N_{\mathrm{s}}\}$ is a set of generated sample points corresponding to $\bm{v}$ and $\bm{h}$. The accuracy of this approximation depends on the sampling quality, that is, how closely the sample distribution follows the target distribution. The sampling quality can be improved by using an initial distribution close to the target distribution. The CD and PCD methods~\cite{Hinton2002_rbm,Tieleman2008_PCD}, which are well-known learning methods for RBMs, set the data distribution and the empirical distribution of sample points used to compute the parameter gradient in the previous iteration, respectively, as their initial distributions. These choices of initial distribution are expected to enable high-quality sampling even with only a few transitions. The sampling quality can also be improved by increasing the number of transitions, $s$. However, this increases the computational cost of evaluating the parameter gradients and is therefore undesirable. 

The approximation accuracy can be further improved using spatial Monte Carlo integration (SMCI)~\cite{Yasuda2015_SMCI,Yasuda2021_SMCI}, which reduces the variance of the estimator by properly summing over a subset of variables. SMCI approximates the expectation using a sample average of conditional expectations; for example, the expectation of $v_i$ is approximated by
\begin{align*}
\mathbb{E}_{P_{\theta}(\bm{v}, \bm{h})}[v_i] 
&= \sum_{\bm{h}} \left(\sum_{\bm{v}} v_i P_\theta(\bm{v}\mid\bm{h})\right) P_\theta^{(h)}(\bm{h}) \\
&= \sum_{\bm{h}} \sigmoid\bigl(\eta_i(\bm{h};\theta)\bigr) P_\theta^{(h)}(\bm{h}) \\
&\approx \frac{1}{N_{\mathrm{s}}}\sum_{\nu=1}^{N_{\mathrm{s}}} \sigmoid\bigl(\eta_i(\mathbf{h}^{(\nu)};\theta)\bigr),
\end{align*}
where $P_{\theta}^{(h)}(\bm{h})$ is the marginal distribution for the hidden variables. SMCI-based learning achieved a higher log-likelihood than learning based on standard Monte Carlo integration, where the expectation is approximated by a sample average~\cite{Sekimoto2023}. However, under some learning settings, the SMCI-based learning method exhibited unexpected behavior, in which the log-likelihood decreased substantially. 

\subsection{Preliminary Investigation of Learning Failure} \label{ssec:demonstrate_failure}

In this section, we numerically identify the learning settings in which SMCI-based learning fails and investigate the cause of this failure. The training dataset was generated from a generative RBM with $n=10$ visible and $m=100$ hidden variables. Its parameters were independently drawn from a normal distribution whose mean and variance were zero and $0.01$, respectively. We then generated $N_{\mathrm{d}}=100$ training data points from the generative RBM by running BGS, where the initial distribution was set to the uniform distribution and the number of steps was $5000$. We referred to this dataset as the \textit{GenRBM} dataset. Next, we constructed a training RBM with $n=10$ visible and $m=100$ hidden variables, with the bias and connection parameters were initialized to zero and using Gaussian-type Xavier initialization~\cite{Xavier2010}, respectively. The training RBM was then trained using a sampling approximation with a PCD-like persistent chain and semi-second-order SMCI (see Appendix \ref{app:s2-SMCI}). Here, the number of steps and the sample size were set to $s=1$ and $N_{\mathrm{s}}=100$, respectively. Gradient ascent was performed using AdaMax~\cite{Adam2015_adamax}, with the learning rate fixed at $0.01$ and the other hyperparameters set to their default values.

We monitored the learning behavior using three quantities: the log-likelihood in Eq.~\eqref{eq:log_likelihood}, the mean absolute errors (MAEs) for the expectation approximation during the learning, and the second-largest eigenvalue of the BGS transition matrix. A higher log-likelihood indicates better learning performance with less influence from approximating the expectations. Since the number of visible variables $n$ is small in this experiment setting, the probability value for the visible distribution $P_\theta^{(v)}(\bm{v})$ can be exactly computed without using any approximation; thus, the log-likelihood can be evaluated. Furthermore, the expectation over the visible distribution $P_\theta^{(v)}(\bm{v})$ can also be exactly computed, and the expectations evaluated during training, $\mathbb{E}_{P_{\theta}(\bm{v}, \bm{h})}[v_i]$, $\mathbb{E}_{P_{\theta}(\bm{v}, \bm{h})}[h_j]$, and $ \mathbb{E}_{P_{\theta}(\bm{v}, \bm{h})}[v_i h_j]$, can thus be computed by rewriting $\mathbb{E}_{P_{\theta}(\bm{v}, \bm{h})}[v_i]$ and $\mathbb{E}_{P_{\theta}(\bm{v}, \bm{h})}[h_j]$ as 
\begin{align*}
\mathbb{E}_{P_{\theta}(\bm{v}, \bm{h})}[h_j] &= \mathbb{E}_{P_\theta^{(v)}(\bm{v})}[\sigmoid(\zeta_j(\bm{v};\theta))], \\
\mathbb{E}_{P_{\theta}(\bm{v}, \bm{h})}[v_i h_j] &= \mathbb{E}_{P_\theta^{(v)}(\bm{v})}[v_i \sigmoid(\zeta_j(\bm{v};\theta))],
\end{align*}
respectively. Let $\mathcal{M}_i$, $\mathcal{M}_j$, and $\mathcal{M}_{i,j}$ denote the estimators used during learning for these three expectations, and their MAEs are then written as
\begin{align*}
\mathrm{MAE}_b &:= \frac{1}{n}\sum_{i=1}^{n} \bigl| \mathbb{E}_{P_{\theta}(\bm{v}, \bm{h})}[v_i] - \mathcal{M}_i \bigr|, \\
\mathrm{MAE}_c &:= \frac{1}{m}\sum_{j=1}^{m} \bigl| \mathbb{E}_{P_{\theta}(\bm{v}, \bm{h})}[h_j] - \mathcal{M}_j \bigr|, \\
\mathrm{MAE}_W &:= \frac{1}{nm}\sum_{i=1}^{n} \sum_{j=1}^{m} \bigl| \mathbb{E}_{P_{\theta}(\bm{v}, \bm{h})}[v_i h_j] - \mathcal{M}_{i,j} \bigr|.
\end{align*}
The MAEs directly measure the approximation accuracy of the expectations during learning. Larger MAEs indicate less accurate approximations. When the second-largest eigenvalue $\lambda_2$ is close to one, the relaxation time is long. Therefore, a large $\lambda_2$ means that BGS strongly depends on the initial state and requires a large number of steps to obtain high-quality samples. The method used to evaluate $\lambda_2$ used in this experiment is described in Appendix~\ref{app:trans_matrix}. 

\begin{figure}
    \centering
    \includegraphics[width=\linewidth]{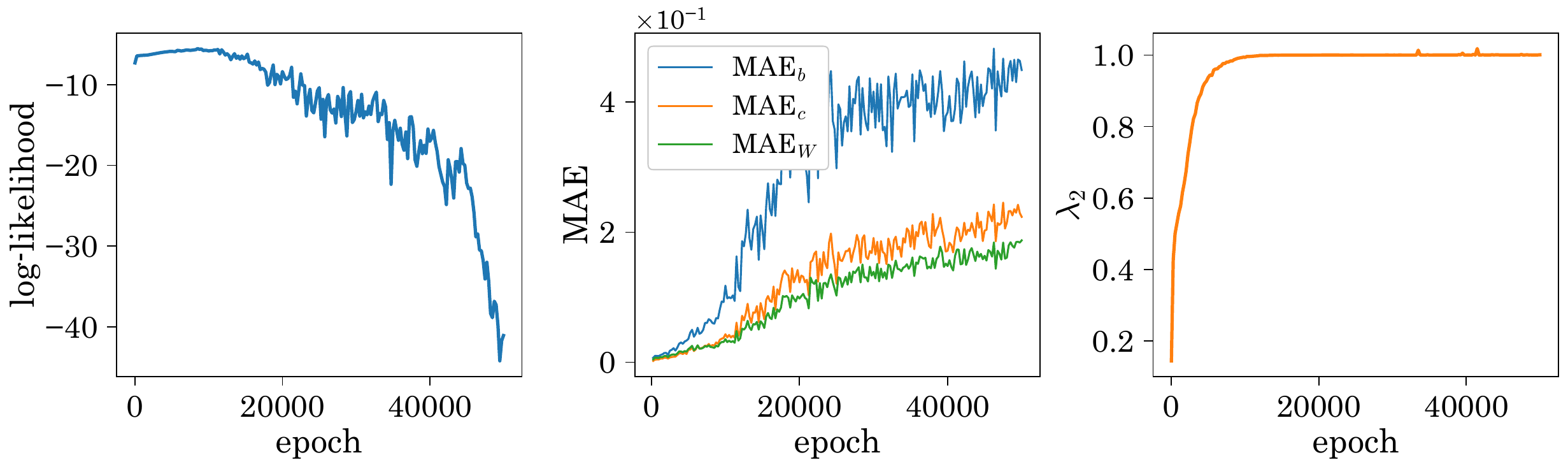}
    \caption{Log-likelihood, the MAEs, and the second-largest eigenvalue $\lambda_2$ versus the training epoch for the GenRBM dataset. The results represent averages over 10 experiments.}
    \label{fig:learning_failure}
\end{figure}

Figure~\ref{fig:learning_failure} shows three quantities characterizing the learning behavior as functions of the training epoch. As training progressed, the MAEs increased, implying that the parameter gradients were estimated less accurately. This deterioration in gradient estimation, in turn, led to a decrease in the log-likelihood. Moreover, the estimator of $\lambda_2$ converged to one as the training began to fail. As the relaxation time increased, the sampling became more strongly dependent on the initial states, making it more difficult to sample from the target distribution within a limited number of sampling steps. These results suggest that the resulting degradation in the sampling quality of BGS caused the learning process to fail. BGS is a local transition kernel, and increasing the relaxation time implies that its locality becomes stronger. Thus, the learning failure could be mitigated by introducing transition kernels that enable nonlocal transitions in the visible state space.

\section{Deep Tempering} \label{sec:DT}

\begin{figure}
    \centering
    \includegraphics[width=0.4\linewidth]{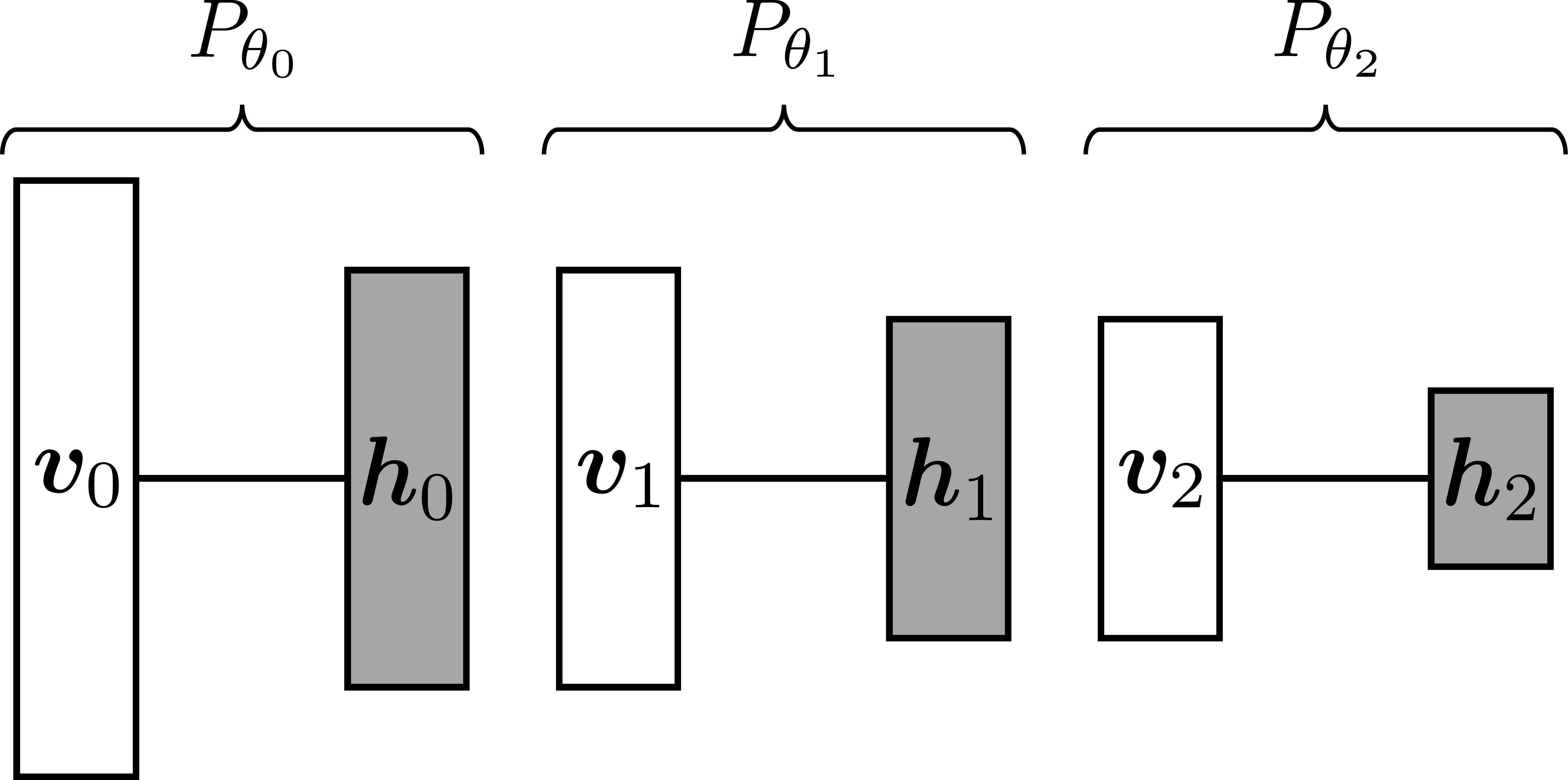}
    \caption{Illustration of the RBM sequence with $L=2$ used in DT. The bottommost RBM corresponds to the training RBM, and the other RBMs are used for sampling.}
    \label{fig:DeepTempering}
\end{figure}

DT~\cite{desjardins2014_deeptempering} is a global transition kernel based on parallel tempering~\cite{swendsen1986_ReMC,hukushima1996_ReMC} for RBMs. A sequence of distributions for parallel tempering is obtained by stacking auxiliary RBMs and learning them jointly with the training RBM. The training RBM is denoted by $P_{\theta_0}(\bm{v}_{0}, \bm{h}_{0})$, and we consider $L$ additional RBMs expressed as
\begin{align*}
P_{\theta_\ell}(\bm{v}_{\ell}, \bm{h}_{\ell}). \qquad(\ell=1,2,\ldots,L)
\end{align*}
Here, $\theta_{\ell}$ denotes the parameter of the $\ell$-th RBM, and $\bm{v}_{\ell} \in \{0,1\}^{n_{\ell}}$ and $\bm{h}_{\ell} \in \{0,1\}^{m_{\ell}}$ are the visible and hidden variables for $\ell$-th RBM. The parameters are collectively denoted by $\Theta := (\theta_0,\theta_1,\ldots,\theta_L)$. Here, the number of visible variables, respectively, of the $(\ell+1)$-th RBM is constrained to equal the number of hidden variables in the $\ell$-th RBM, i.e., $n_{\ell+1} = m_{\ell}$ for $\ell=0,1,\ldots,L-1$. Figure~\ref{fig:DeepTempering} illustrates the sequence of the training RBM and the additional RBMs. The training and additional RBMs are trained by iteratively updating the parameters as
\begin{align}
\Theta^{(t+1)} \leftarrow \Theta^{(t)} + \varepsilon \nabla_{\Theta'} \bar{\ell}(\Theta'\mid\Theta^{(t)})\big|_{\Theta'=\Theta^{(t)}},
\end{align}
using the function defined by
\begin{align}
\bar{\ell}(\Theta'\mid\Theta) := \ell(\theta_0') + \sum_{\ell=1}^{L} \mathbb{E}_{Q_{\mathfrak{D}}^{(\ell)}(\bm{v}_{\ell}\mid\Theta)}[\ln P_{\theta_{\ell}'}(\bm{v}_{\ell})],
\end{align}
where $Q_{\mathfrak{D}}^{(\ell)}$ denotes the extended data distribution represented as
\begin{align}
Q_{\mathfrak{D}}^{(\ell)}(\bm{v}_{\ell}\mid\Theta) :=  \sum_{\bm{v}_{\ell-1}} \sum_{\bm{h}_{\ell-1}} \delta(\bm{v}_{\ell}, \bm{h}_{\ell-1}) P_{\theta_{\ell-1}}(\bm{h}_{\ell-1}\mid \bm{v}_{\ell-1}) Q_{\mathfrak{D}}^{(\ell)}(\bm{v}_{\ell-1}\mid\Theta),
\label{eq:extended_data_dist}
\end{align}
where $Q_{\mathfrak{D}}^{(0)} = Q_{\mathfrak{D}}$. The gradient with respect to the $\ell$-th parameter $\theta_\ell$ is written as
\begin{align}
\nabla_{\theta_{\ell}'} \bar{\ell}(\Theta'\mid\Theta) = - \mathbb{E}_{Q_{\mathfrak{D}}^{(\ell)}(\bm{v}_{\ell}\mid\Theta)}[\nabla_{\theta'_\ell} \hat{E}(\bm{v}_\ell;\theta'_\ell)]  + \mathbb{E}_{P_{\theta_\ell}(\bm{v}_\ell, \bm{h}_\ell)}\bigl[\nabla_{\theta'_\ell} E(\bm{v}_\ell, \bm{h}_\ell;\theta'_\ell)\bigr].
\label{eq:grad_Deep}
\end{align}
The first term can be readily estimated using a sampling approximation based on ancestral sampling from the extended data distribution in Eq.~\eqref{eq:extended_data_dist}. In contrast, the second term is estimated using a sampling approximation based on parallel tempering on the joint distribution $P_{\Theta}(\bm{V}) := \prod_{\ell=0}^{L} P_{\theta_{\ell}}^{(v)}(\bm{v}_{\ell})$, where $\bm{V}:=(\bm{v}_0, \bm{v}_1,\ldots,\bm{v}_{L})$.

\begin{algorithm}[t]
\caption{Swap kernel}
\label{alg:swap_kernel}
\begin{algorithmic}[1]
\Function{SwapKernel}{$\ell,\mathbf{h}_\ell,\mathbf{v}_{\ell+1}$}
\State Sample $\mathrm{u}\sim\operatorname{Unif}(0,1)$
\If{$\mathrm{u}\leq a_{\ell,\ell+1}(\mathbf{h}_\ell,\mathbf{v}_{\ell+1})$}
\State $(\mathbf{h}_\ell',\mathbf{v}_{\ell+1}')\gets(\mathbf{v}_{\ell+1},\mathbf{h}_\ell)$
\Else
\State $(\mathbf{h}_\ell',\mathbf{v}_{\ell+1}')\gets(\mathbf{h}_\ell,\mathbf{v}_{\ell+1})$
\EndIf
\State \Return $(\mathbf{h}_\ell',\mathbf{v}_{\ell+1}')$
\EndFunction
\end{algorithmic}
\end{algorithm}

DT considers the swap probability between the $\ell$-th hidden variables and the $(\ell+1)$-th visible variables, defined by
\begin{align}
a_{\ell,\ell+1}(\bm{h}_\ell,\bm{v}_{\ell+1}) := \min\left(1,\frac{P_{\theta_{\ell+1}}^{(v)}(\bm{h}_\ell) P_{\theta_{\ell}}^{(h)}(\bm{v}_{\ell+1})}{P_{\theta_{\ell+1}}^{(v)}(\bm{v}_{\ell+1}) P_{\theta_{\ell}}^{(h)}(\bm{h}_{\ell})}\right).
\label{eq:swap_prob}
\end{align}
The corresponding swap kernel is defined by
\begin{align}
\begin{split}
S_{\ell,\ell+1}(\bm{h}_\ell',\bm{v}_{\ell+1}'\mid\bm{h}_\ell,\bm{v}_{\ell+1}) 
&:= a_{\ell,\ell+1}(\bm{h}_\ell,\bm{v}_{\ell+1}) \delta(\bm{h}_\ell', \bm{v}_{\ell+1}) \delta(\bm{v}_{\ell+1}', \bm{h}_\ell) \\
&\qquad+ \bigl[1 - a_{\ell,\ell+1}(\bm{h}_\ell,\bm{v}_{\ell+1})\bigr] \delta(\bm{v}_{\ell+1}', \bm{v}_{\ell+1}) \delta( \bm{h}_\ell', \bm{h}_\ell),
\end{split}
\label{eq:swap_kernel}
\end{align}
and satisfies the detailed balance condition with respect to $P_{\theta_{\ell+1}}^{(v)}(\bm{v}_{\ell+1}) P_{\theta_{\ell}}^{(h)}(\bm{h}_{\ell}) $. The implementation of the swap kernel is described by Algorithm~\ref{alg:swap_kernel}. Using the swap kernel, we define a visible-space swap kernel as
\begin{align}
\bar{K}_{\ell,\ell+1}(\bm{v}_{\ell}',\bm{v}_{\ell+1}'\mid\bm{v}_{\ell},\bm{v}_{\ell+1}) := \sum_{\bm{h}_{\ell}} \sum_{\bm{h}_{\ell}'} P_{\theta_\ell}(\bm{v}_{\ell}' \mid \bm{h}_{\ell}') S_{\ell,\ell+1}(\bm{h}_\ell',\bm{v}_{\ell+1}'\mid\bm{h}_\ell,\bm{v}_{\ell+1}) P_{\theta_\ell}(\bm{h}_{\ell}\mid\bm{v}_{\ell}).
\label{eq:swap_kernel_vis}
\end{align}
This satisfies the balance condition for $P_{\theta_{\ell}}^{(v)}(\bm{v}_{\ell}) P_{\theta_{\ell+1}}^{(v)}(\bm{v}_{\ell+1})$. DT performs parallel tempering using the visible-space swap kernel in Eq.~\eqref{eq:swap_kernel_vis} and BGS on each RBM (see Algorithm~\ref{alg:deep_tempering}). The corresponding transition kernel is denoted by $T_{\Theta}^{\mathrm{DT}}(\bm{V}'\mid\bm{V})$. For simplicity, the parity indicator is omitted from the notation. At each transition, the parity indicator is passed to the next transition and is deterministically updated according to $c'=1-c$. The transition kernel satisfies the balance condition with respect to $P_{\Theta}(\bm{V})$; that is, it leaves $P_{\Theta}(\bm{V})$ invariant~\cite{desjardins2014_deeptempering}. We consider a sample distribution based on DT expressed as
\begin{align*}
P_{\mathrm{DT}}^{s}(\bm{V},\bm{H}) := \prod_{\ell=0}^{L} P_{\theta_{\ell}}(\bm{h}_{\ell}\mid\bm{v}_{\ell}) \sum_{\bm{V}'}T_{\Theta,s}^{\mathrm{DT}}(\bm{V}\mid\bm{V}') P_0(\bm{V}'),
\end{align*}
where $P_0$ and $T_{\Theta,s}^{\mathrm{DT}}$ denote the initial distribution and the $s$-step transition kernel of DT, respectively. Here, $\bm{H}:=(\bm{h}_0, \bm{h}_1,\ldots,\bm{h}_{L})$. Using sample points generated from the sample distribution, the expectation in the second term of the parameter gradient in Eq.~\eqref{eq:grad_Deep} is approximated by a sample average.

\begin{algorithm}[t]
\caption{One-step transition of deep tempering}
\label{alg:deep_tempering}
\begin{algorithmic}[1]
\Require Current visible states $\mathbf{V}=(\mathbf{v}_0,\ldots,\mathbf{v}_L)$ and parity indicator $c\in\{0,1\}$
\Ensure Updated visible states $\mathbf{V}'=(\mathbf{v}_0',\ldots,\mathbf{v}_L')$ and parity indicator $c'$
\State $\tilde{\mathbf{V}}\gets\mathbf{V}$, \quad $\mathcal{I}_c\gets\{\ell\in\{0,\ldots,L-1\}\mid \ell\equiv c\pmod{2}\}$
\ForAll{$\ell\in\mathcal{I}_c$} \Comment{Visible-space swap}
\State Sample $\mathbf{h}_\ell\sim P_{\theta_\ell}(\bm{h}_\ell\mid\tilde{\mathbf{v}}_\ell)$
\State $(\tilde{\mathbf{h}}_\ell,\tilde{\mathbf{v}}_{\ell+1})\gets\Call{SwapKernel}{\ell,\mathbf{h}_\ell,\tilde{\mathbf{v}}_{\ell+1}}$ 
\State Sample $\tilde{\mathbf{v}}_\ell\sim P_{\theta_\ell}(\bm{v}_\ell\mid\tilde{\mathbf{h}}_\ell)$
\EndFor
\ForAll{$\ell\in\{0,\ldots,L\}$} \Comment{One-step BGS}
\State Sample $\mathbf{h}_\ell'\sim P_{\theta_\ell}(\bm{h}_\ell\mid\tilde{\mathbf{v}}_\ell)$ and $\mathbf{v}_\ell'\sim P_{\theta_\ell}(\bm{v}_\ell\mid\mathbf{h}_\ell')$
\EndFor
\State $c'\gets1-c$
\State \Return $(\mathbf{V}',c')$
\end{algorithmic}
\end{algorithm}

In standard parallel tempering, a sequence of intermediate distributions is constructed by varying an inverse-temperature ladder. In contrast, DT constructs the sequence through learning. This learning-based construction can reduce the number of intermediate distributions required for effective sampling. Indeed, DT has been shown to achieve competitive or better empirical performance than standard parallel tempering while using fewer intermediate distributions~\cite{desjardins2014_deeptempering}. Moreover, because higher-layer RBMs typically have fewer hidden units than lower-layer RBMs, their state spaces are smaller, which is expected to facilitate mixing. However, the swap mechanism, which plays a key role in enabling a nonlocal transition in the visible state space of a lower RBM, is local with respect to the RBM index $\ell$, and a single transition based on DT can only moves a state by at most one layer. Therefore, for a state at the $\ell'$-th RBM to reach the $\ell$-th RBM, where $\ell'>\ell$, at least $(\ell'-\ell)$ DT transitions are required. Thus, although DT can in principle achieve a nonlocal transition in the visible state space of the training RBM, such a transition necessarily requires multiple DT transitions. To address this limitation, we propose a transition kernel that can transfer a state directly between nonadjacent RBMs in a single transition. The proposed kernel thereby enables a nonlocal transition within one step while leaving the target distribution invariant. 

\section{Proposed Non-local Transition Kernel} \label{sec:proposed}

\begin{figure}[t]
  \centering
  \includegraphics[width=0.6\linewidth]{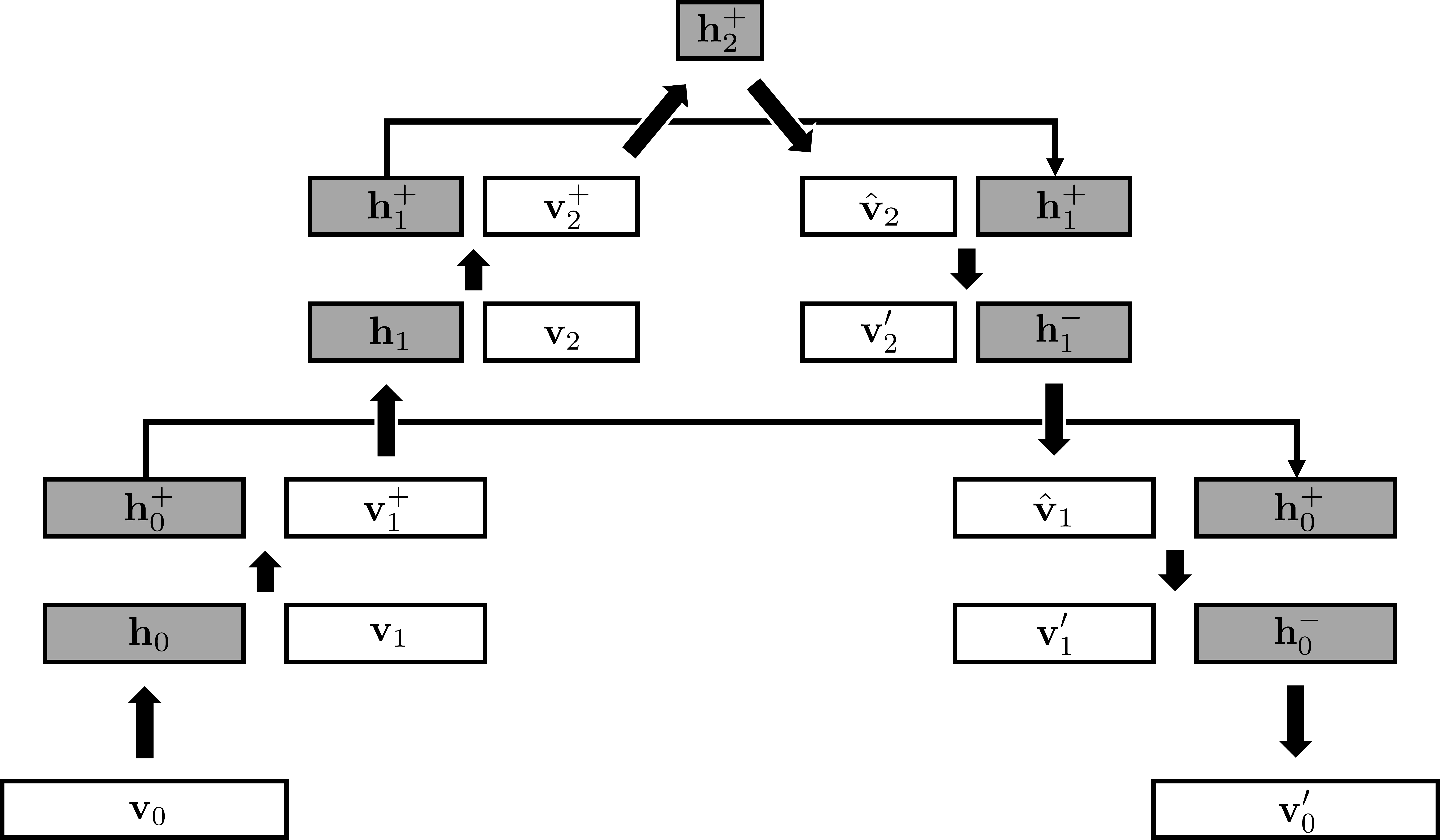}
  \caption{Illustration of the sampling procedure for the proposed transition kernel on an RBM sequence with $L=2$. The upward pass sequentially applies the swap operations from lower to upper RBMs, followed by a one-step BGS transition at the topmost RBM, and the downward pass applies the swap operations in the reverse order of the upward pass.}
  \label{fig:proposed_sampling}
\end{figure}

The proposed transition kernel is formulated recursively and has a round-trip structure: the states are first propagated upward through the RBM sequence, one BGS transition is performed at the topmost RBM, and the resulting state is subsequently propagated downward. For the topmost RBM, we use the one-step BGS kernel as the base transition kernel $K_L$, i.e., $K_L(\bm{v}_L'\mid\bm{v}_L)
:=T_{\theta_L}^{\mathrm{BGS}}(\bm{v}_L'\mid\bm{v}_L)$. We next consider the topmost RBM and the RBM immediately below it. Starting from the current state $(\mathbf{v}_{L-1},\mathbf{v}_L)$, we first sample $\mathbf{h}_{L-1}\sim P_{\theta_{L-1}}(\bm{h}_{L-1}\mid\mathbf{v}_{L-1})$. The sampled state $\mathbf{h}_{L-1}$ and the current state $\mathbf{v}_L$ are then updated using the swap kernel $S_{L-1,L}$, yielding the post-swap states $(\mathbf{h}_{L-1}^+,\mathbf{v}_L^+)$. The kernel $T_L$ is subsequently applied to $\mathbf{v}_L^+$, producing the intermediate state $\hat{\mathbf{v}}_L$. During the downward part of the transition, $\mathbf{h}_{L-1}^+$ and $\hat{\mathbf{v}}_L$ are updated again using $S_{L-1,L}$, yielding $(\mathbf{h}_{L-1}^-,\mathbf{v}_L')$. Finally, $\mathbf{v}_{L-1}'$ is sampled from $P_{\theta_{L-1}}(\bm{v}_{L-1}\mid\mathbf{h}_{L-1}^-)$. The corresponding transition kernel is defined by
\begin{align}
\begin{split}
T_{L-1}(\bm{v}_{L-1}',\bm{v}_L'\mid\bm{v}_{L-1},\bm{v}_L)
&:=\sum_{\bm{h}_{L-1}^{-},\,\bm{h}_{L-1}^{+},\,\hat{\bm{v}}_L,\,\bm{v}_L^{+},\,\bm{h}_{L-1}}
P_{\theta_{L-1}}(\bm{v}_{L-1}'\mid\bm{h}_{L-1}^{-}) \\
&\qquad\times S_{L-1,L}(\bm{h}_{L-1}^{-},\bm{v}_L'\mid\bm{h}_{L-1}^{+},\hat{\bm{v}}_L)
T_L(\hat{\bm{v}}_L\mid\bm{v}_L^{+}) \\
&\qquad\times S_{L-1,L}(\bm{h}_{L-1}^{+},\bm{v}_L^{+}\mid\bm{h}_{L-1},\bm{v}_L)
P_{\theta_{L-1}}(\bm{h}_{L-1}\mid\bm{v}_{L-1}).
\end{split}
\label{eq:Lminus1_transition}
\end{align}
We next consider a more general transition kernel defined on the visible variables from the $\ell$-th RBM to the topmost RBM. We denote the corresponding collection of visible variables by $\bm{V}_{\ell:L}:=(\bm{v}_\ell,\bm{v}_{\ell+1},\ldots,\bm{v}_L)$ . Given the state $\mathbf{V}_{\ell:L}$, we first sample $\mathbf{h}_\ell\sim P_{\theta_\ell}(\bm{h}_\ell\mid\mathbf{v}_\ell)$. The sampled state $\mathbf{h}_\ell$ and the current state $\mathbf{v}_{\ell+1}$ are updated using $S_{\ell,\ell+1}$, producing $(\mathbf{h}_\ell^+,\mathbf{v}_{\ell+1}^+)$. The recursively defined kernel $T_{\ell+1}$ is then applied to $(\mathbf{v}_{\ell+1}^+,\mathbf{V}_{\ell+2:L})$, yielding $(\hat{\mathbf{v}}_{\ell+1},\mathbf{V}_{\ell+2:L}')$. Here, $\hat{\mathbf{v}}_{\ell+1}$ denotes the intermediate visible state returned to the $(\ell+1)$-th RBM after completing the round trip through the upper sequence. During the downward transition, $\mathbf{h}_\ell^+$ and $\hat{\mathbf{v}}_{\ell+1}$ are updated using $S_{\ell,\ell+1}$, producing $(\mathbf{h}_\ell^-,\mathbf{v}_{\ell+1}')$. Finally, $\mathbf{v}_\ell'$ is sampled from $P_{\theta_\ell}(\bm{v}_\ell\mid\mathbf{h}_\ell^-)$. Accordingly, for $\ell=0,1,\ldots,L-1$, the recursive transition kernel is defined as
\begin{align}
\begin{split}
T_{\ell}(\bm{V}_{\ell:L}'\mid\bm{V}_{\ell:L})
&:=\sum_{\bm{h}_{\ell}^{-},\,\bm{h}_{\ell}^{+},\,\hat{\bm{v}}_{\ell+1},\,\bm{v}_{\ell+1}^{+},\,\bm{h}_{\ell}}
P_{\theta_{\ell}}(\bm{v}_{\ell}'\mid\bm{h}_{\ell}^{-}) \\
&\qquad\times S_{\ell,\ell+1}(\bm{h}_{\ell}^{-},\bm{v}_{\ell+1}'\mid\bm{h}_{\ell}^{+},\hat{\bm{v}}_{\ell+1}) \\
&\qquad\times T_{\ell+1}(\hat{\bm{v}}_{\ell+1},\bm{V}_{\ell+2:L}'\mid\bm{v}_{\ell+1}^{+},\bm{V}_{\ell+2:L}) \\
&\qquad\times S_{\ell,\ell+1}(\bm{h}_{\ell}^{+},\bm{v}_{\ell+1}^{+}\mid\bm{h}_{\ell},\bm{v}_{\ell+1})
P_{\theta_{\ell}}(\bm{h}_{\ell}\mid\bm{v}_{\ell}).
\end{split}
\label{eq:ell_trans}
\end{align}
The proposed transition kernel for the entire RBM sequence is obtained by setting $\ell=0$, i.e., 
\begin{align*}
T_{\Theta}^{\mathrm{prop}}(\bm{V}'\mid\bm{V}) := T_0(\bm{V}'\mid\bm{V}).
\end{align*}
Figure~\ref{fig:proposed_sampling} illustrates the round-trip sampling procedure for the proposed transition kernel, and Algorithm~\ref{alg:proposed_sampling} shows its implementation. The proposed transition kernel leaves $P_{\Theta}(\bm{V})$ invariant, as described in Appendix~\ref{app:invariant_distribution}. Learning based on the proposed transition kernel follows the same procedure as learning based on DT. Specifically, the parameter set $\Theta$ is updated by gradient ascent using the parameter gradient in Eq.~\eqref{eq:grad_Deep}, and the expectation appearing in the gradient is evaluated using a sampling approximation based on the proposed transition kernel.

Owing to its round-trip structure, the proposed transition kernel may transfer a state from an upper-layer RBM, where mixing is relatively easy, to the bottommost RBM (i.e., the training RBM) within a single transition. Consequently, nonlocal transitions in the visible state space of the training RBM are expected to occur within a small number of transition steps. 

\begin{algorithm}[t]
\caption{One-step transition of the proposed transition kernel}
\label{alg:proposed_sampling}
\begin{algorithmic}[1]
\Require Current visible-state realization $\mathbf{V}=(\mathbf{v}_0,\ldots,\mathbf{v}_L)$
\Ensure Updated visible-state realization $\mathbf{V}'=(\mathbf{v}_0',\ldots,\mathbf{v}_L')$
\State $\mathbf{v}_0^+\gets \mathbf{v}_0$
\For{$\ell=0,\ldots,L-1$} \Comment{Upward pass}
\State Sample $\mathbf{h}_\ell\sim P_{\theta_\ell}(\bm{h}_\ell\mid\mathbf{v}_\ell^+)$
\State $(\mathbf{h}_\ell^+,\mathbf{v}_{\ell+1}^+)\gets\Call{SwapKernel}{\ell,\mathbf{h}_\ell,\mathbf{v}_{\ell+1}}$
\EndFor
\State Sample $\mathbf{h}_L^+\sim P_{\theta_L}(\bm{h}_L\mid\mathbf{v}_L^+)$ and $\hat{\mathbf{v}}_L\sim P_{\theta_L}(\bm{v}_L\mid\mathbf{h}_L^+)$ \Comment{One-step BGS}
\For{$\ell=L-1,\ldots,0$} \Comment{Downward pass}
\State $(\mathbf{h}_\ell^-,\mathbf{v}_{\ell+1}')\gets\Call{SwapKernel}{\ell,\mathbf{h}_\ell^+,\hat{\mathbf{v}}_{\ell+1}}$
\State Sample $\hat{\mathbf{v}}_\ell\sim P_{\theta_\ell}(\bm{v}_\ell\mid\mathbf{h}_\ell^-)$
\EndFor
\State $\mathbf{v}_0'\gets \hat{\mathbf{v}}_0$
\State \Return $\mathbf{V}'$
\end{algorithmic}
\end{algorithm}

\section{Numerical Experiments} \label{sec:num_exp}

In this section, we numerically evaluate the performance of learning using the proposed transition kernel and the nonlocality of its transitions.

\subsection{Experimental Settings} \label{ssec:experimental-settings}

We evaluated the learning performance on six datasets: GenRBM, islands, pentagon, iris, wine, and seeds. The RBM parameter initialization, optimization method, and sampling approximations for the first and second terms of the parameter gradient in Eq.~\eqref{eq:grad_Deep} were common to all datasets. Specifically, for both the training RBM and the stacked RBMs, the bias parameters were initialized to zero, while the weight parameters were initialized using Gaussian Xavier initialization~\cite{Xavier2010}. The parameters were optimized by gradient ascent using AdaMax~\cite{Adam2015_adamax}, with its default hyperparameter values except for the learning rate. The expectation in the first term of the parameter gradient was approximated by drawing one sample from the extended data distribution in Eq.~\eqref{eq:extended_data_dist} for each data point. Thus, the total sample size was equal to the dataset size. In contract, the expectation in the second term was approximated using PCD-like persistent chains and semi-second-order SMCI, with the number of transition steps set to $s=1$. For DT and the proposed transition kernel, we used an RBM sequence with $L=3$ for all datasets unless otherwise specified. The numbers of hidden variables in the RBMs stacked above the training RBM were set to $(m_1,m_2,m_3)=(50,25,12)$. The remaining experimental settings are described below, and the dataset-specific hyperparameters are summarized in Table~\ref{tab:experimental-settings}. 

\begin{table}[t]
  \centering
  \caption{Experimental settings for training the RBMs.}
  \label{tab:experimental-settings}
  \small
  \setlength{\tabcolsep}{4pt}
  \begin{tabular}{lcccccc}
    \toprule
    Setting & GenRBM & Islands & Pentagon & Iris & Wine & Seeds \\
    \midrule
    Data dimension $n$ & 10 & 25 & 25 & 20 & 65 & 35 \\
    Data size $N_{\mathrm{d}}$ & 100 & 100 & 100 & 150 & 178 & 210 \\
    Number of hidden variables $m$ & 100 & 100 & 100 & 100 & 100 & 100 \\
    Sample size $N_{\mathrm{s}}$ & 100 & 100 & 100 & 150 & 100 & 100 \\
    Learning rate $\varepsilon$ & 0.01 & 0.001 & 0.001 & 0.002 & 0.001 & 0.001 \\
    \bottomrule
  \end{tabular}
\end{table}

\paragraph{GenRBM Dataset.}
The GenRBM dataset used in the preliminary experiment was also used in the main experiment. For details of this dataset, see Section~\ref{ssec:demonstrate_failure}. For DT and the proposed transition kernel, in addition to the default setting of $L=3$, we examined shallower RBM sequences with $L=1$ and $L=2$. For $L=1$, $2$, and $3$, the numbers of hidden variables in the stacked RBMs were set to $m_1=50$, $(m_1,m_2)=(50,25)$, and $(m_1,m_2,m_3)=(50,25,12)$, respectively.

\begin{figure}[t]
    \centering
    \includegraphics[width=0.55\linewidth]{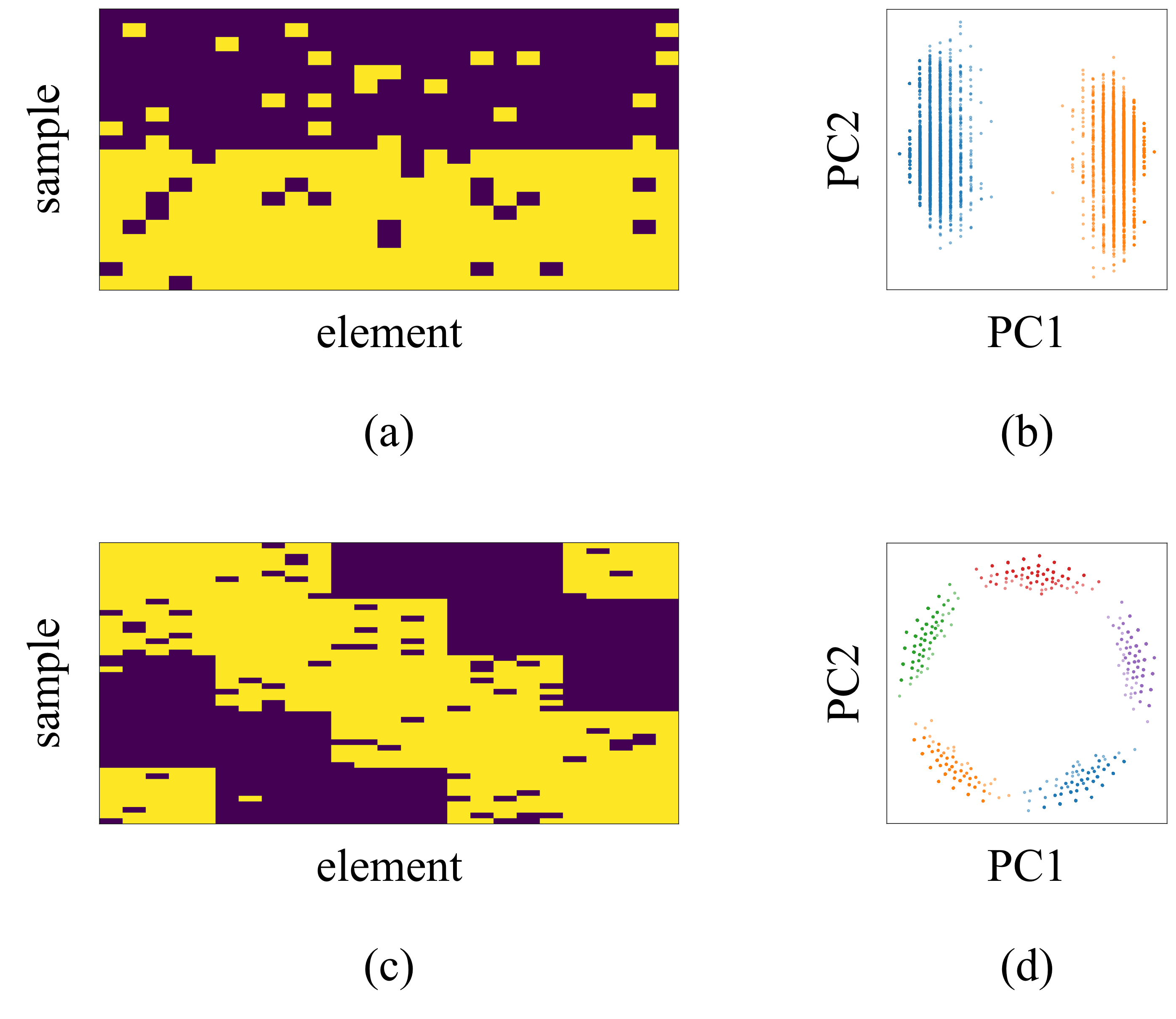}
    \caption{Visualization of low-rank synthetic datasets with data dimension $n=25$: (a) samples and (b) a PCA projection for the islands dataset with $a=0.4$, and (c) samples and (d) a PCA projection for the pentagon dataset with $\kappa=6$.}
    \label{fig:SyntheticDatasets}
\end{figure}

\paragraph{Islands Dataset.}
The islands dataset is generated from a two-component mixture of product Bernoulli distributions expressed as
\begin{align}
P_{\mathrm{islands}}(\bm{v}\mid a) &:= \frac{1}{2} \sum_{z\in\{-1,+1\}} \prod_{i=1}^{n} \mathrm{Bern}\Bigl(v_i \,\Big|\, \frac{1}{2} + a z\Bigr),
\end{align}
where the parameter $a\in[0,1/2]$ controls the separation between the two mixture components: a larger value of $a$ yields more clearly separated modes. Figure~\ref{fig:SyntheticDatasets} shows the visualization of sample points and a principal component analysis (PCA) projection for the datasets with $a=0.4$. We set $a=0.4$.

\paragraph{Pentagon Dataset.}
The pentagon dataset is generated from a five-component mixture of product Bernoulli distributions expressed as
\begin{align*}
P_{\mathrm{pentagon}}(\bm{v}\mid \kappa) &:= \frac{1}{5} \sum_{z=1}^{5} \prod_{i=1}^{n} \mathrm{Bern}(v_i \mid \sigmoid(\kappa \rho_{z,i})),
\end{align*}
where $\rho_{z,i} := \cos\left[\frac{2\pi}{5}(\chi_i - z)\right]$ and $\chi_i := \left\lfloor\frac{5 i}{n} \right\rfloor$. The parameter $\kappa\ge0$ controls the separation among the five mixture components: a larger value of $\kappa$ yields more clearly separated modes. Figure~\ref{fig:SyntheticDatasets} shows the visualization of the datasets with $\kappa=6$. We set $\kappa=6$.

\paragraph{Iris Dataset.}
The iris dataset is a classification dataset comprising 150 data points from three iris species: iris setosa, iris versicolor, and iris virginica. Each data point has four continuous features: sepal length, sepal width, petal length, and petal width. We transformed each continuous feature into binary variables using rank-hot encoding, also known as thermometer encoding. Specifically, each feature was divided into $K_{\mathrm{enc}}$ bins based on its quantiles, and the bins were ordered by feature value. For $k\in 1,2,\ldots,K_{\mathrm{enc}}$, the $k$-th bin was encoded as a binary vector of dimension $(K_{\mathrm{enc}}-1)$, whose first $(k-1)$ elements were ones and whose remaining elements were zeros. Consequently, the four continuous features were represented by a total of $n=4(K_{\mathrm{enc}}-1)$ binary variables. We set $K_{\mathrm{enc}}=6$, yielding $n=20$.

\paragraph{Wine Dataset.}
The wine dataset is a classification dataset comprising 178 data points from three wine cultivars. Each data point has 13 continuous features obtained from chemical analyses of wine samples. Each continuous feature was transformed into binary variables using rank-hot encoding with $K_{\mathrm{enc}}=6$, in the same manner as for the iris dataset. Consequently, the total data dimension was $n=13(K_{\mathrm{enc}}-1)=65$.

\paragraph{Seeds Dataset.}
The seeds dataset is a classification dataset comprising 210 data points from three wheat varieties. Each data point has seven continuous features describing the geometrical properties of a wheat kernel. Each continuous feature was transformed into binary variables using rank-hot encoding with $K_{\mathrm{enc}}=6$, in the same manner as for the iris and wine datasets. Consequently, the total data dimension was $n=7(K_{\mathrm{enc}}-1)=35$.

\subsection{Learning Performance}

\begin{figure}[t]
    \centering
    \includegraphics[width=0.85\linewidth]{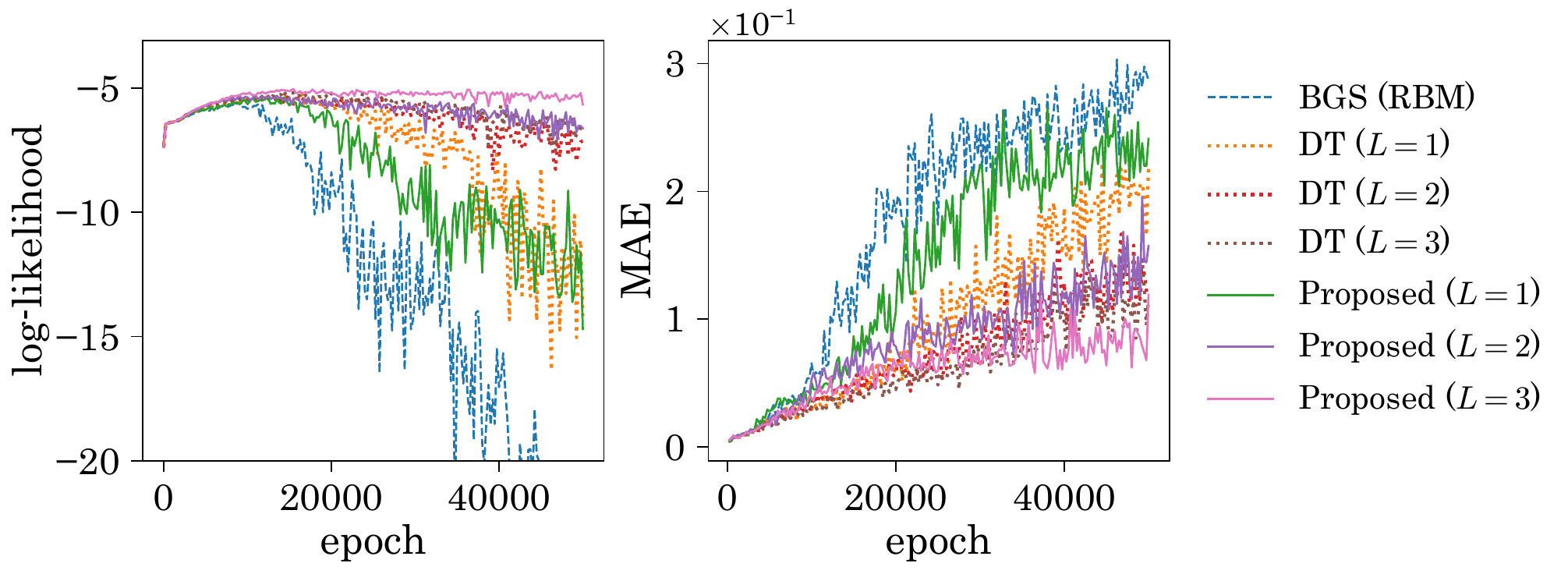}
    \caption{Log-likelihood and MAE versus the training epoch for the GenRBM dataset. The results represent averages over 10 experiments.}
    \label{fig:GenRBM_log_likelihood_mae}
\end{figure}

We first demonstrate that the proposed transition kernel mitigates the learning failure observed for the GenRBM dataset in Section~\ref{ssec:demonstrate_failure}. Figure~\ref{fig:GenRBM_log_likelihood_mae} shows the log-likelihood and MAE as functions of the training epoch for BGS, DT, and the proposed transition kernel. For both DT and the proposed transition kernel, the degradation in the log-likelihood is progressively mitigated as $L$ increases. In particular, the proposed transition kernel mitigates this degradation more effectively than DT, and learning appears stable at $L=3$.

\begin{figure}[t]
    \centering
    \includegraphics[width=0.95\linewidth]{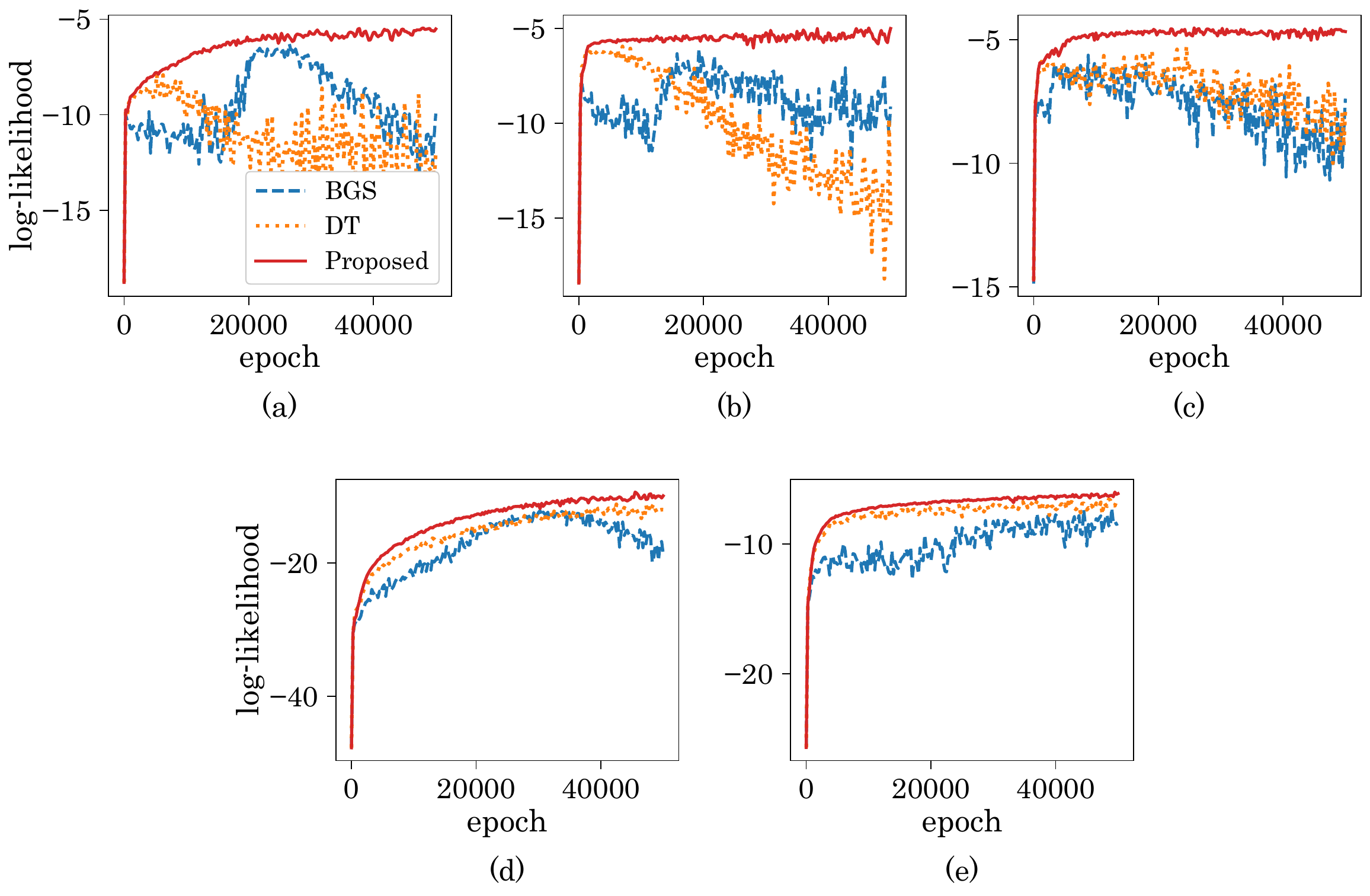}
    \caption{Log-likelihoods versus the training epoch for the (a) islands, (b) pentagon, (c) iris, (d) wine, and (e) seeds datasets. The results represent averages over 10 experiments.}
    \label{fig:cluster_datasets_log_likelihood}
\end{figure}

Next, we compare the learning performance on the five datasets with clustered structures: the islands, pentagon, iris, wine, and seeds datasets. The datasets lead to multimodal learned distributions with high energy barriers, which may make BGS transitions more local. We evaluated the learning performance using the log-likelihood of the training dataset. However, for the wine and seeds datasets, exact computation of the partition function was computationally intractable owing to their high dimensionality. Therefore, we approximated the log-likelihood by estimating the partition function using marginalized annealed importance sampling~\cite{yasuda2022_mAIS}. Figure~\ref{fig:cluster_datasets_log_likelihood} shows the log-likelihood as a function of the training epoch for the five datasets. The proposed transition kernel achieves learning performance comparable to or better than that of BGS and DT across these datasets.

\subsection{Sampling Performance}

We evaluated the sampling quality of the proposed transition kernel using the RBM sequence that achieved the highest log-likelihood among all RBM sequences obtained from the experimental trials. All RBM sequences were trained using the proposed transition kernel. For the GenRBM dataset, the model was selected from those trained with $L=3$.

\begin{figure}[t]
    \centering
    \includegraphics[width=\linewidth]{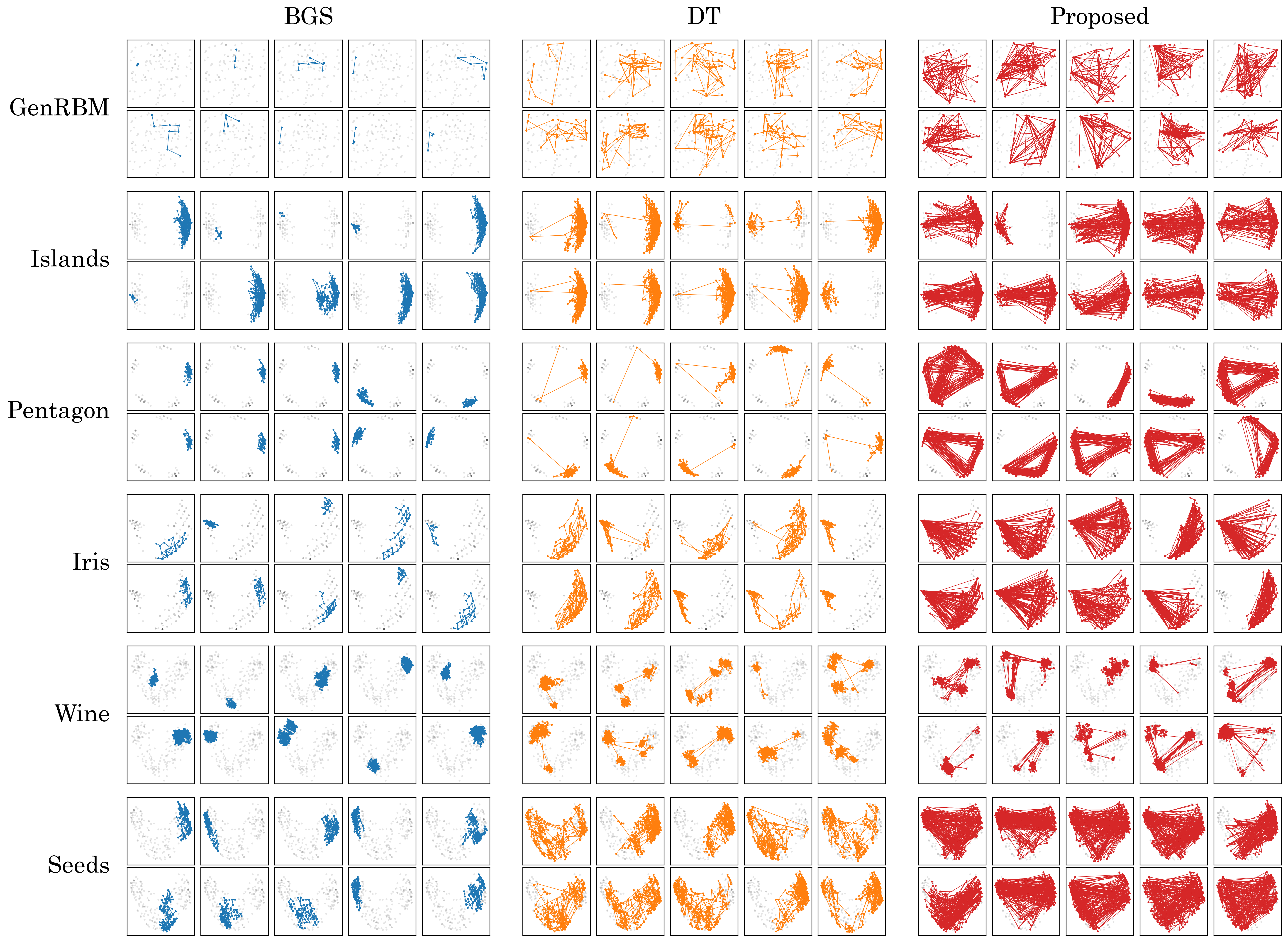}
    \caption{Transition trajectories of BGS, DT, and the proposed kernel for the six datasets.}
    \label{fig:trajectories}
\end{figure}

We visualized the transition trajectories to examine whether nonlocal transitions were achieved. The initial states were set to the states obtained from the persistent chains during training. DT and the proposed transition kernel were then applied to the RBM sequence, and BGS was applied to the bottommost training RBM. We performed $100$ transition steps and plotted the resulting trajectories in the PCA space of the training dataset. Figure~\ref{fig:trajectories} depicts the resulting trajectories for all datasets. The trajectories generated by the proposed transition kernel covered a broader range of states than those generated by the other transition kernels. In particular, for datasets with clustered structures, the proposed transition kernel moved between clusters more frequently.

\begin{figure}[t]
    \centering
    \includegraphics[width=0.7\linewidth]{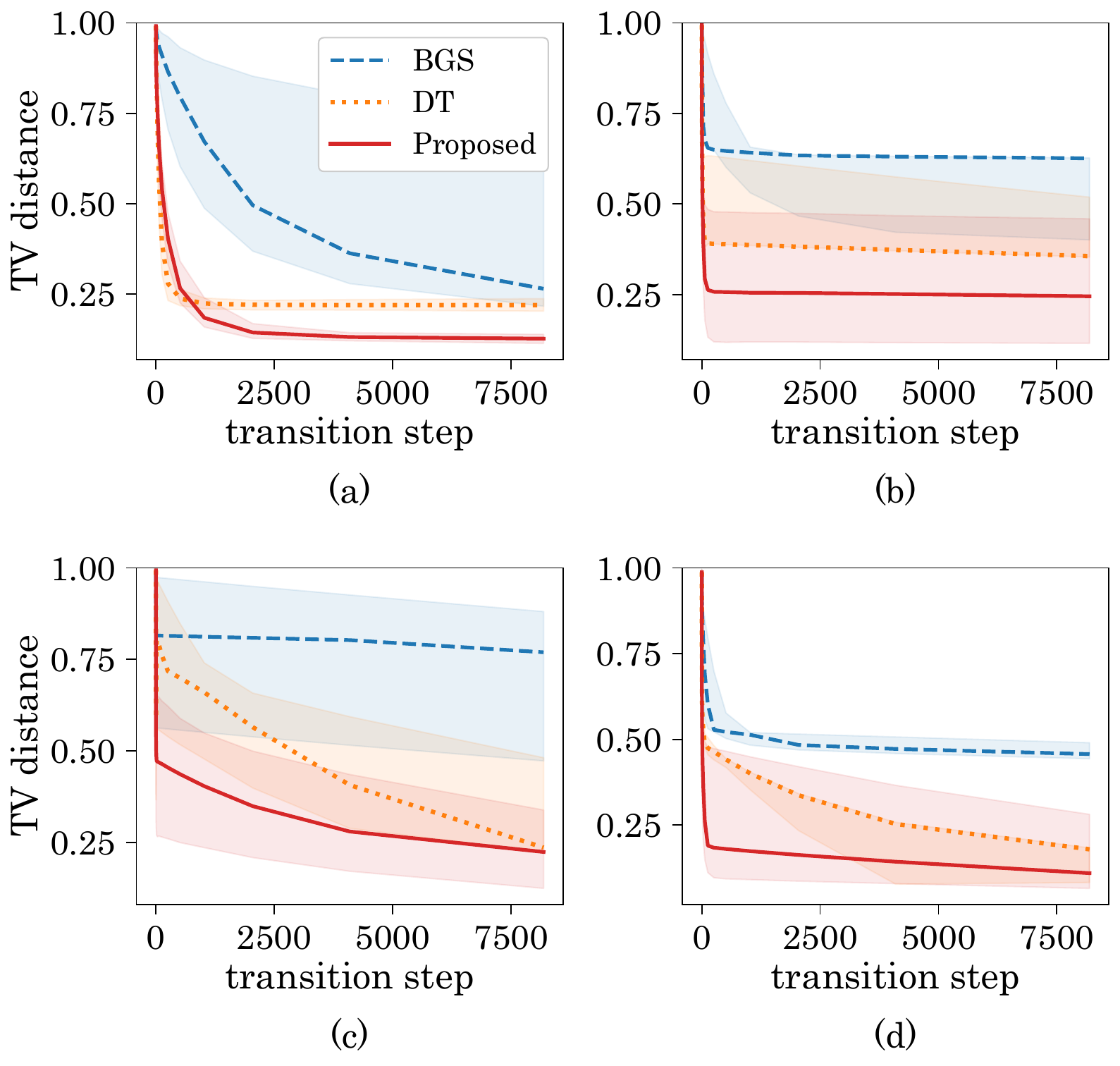}
    \caption{TV distances in Eq.~\eqref{eq:tv_distance} for the (a) GenRBM, (b) islands, (c) pentagon, and (d) iris datasets. The solid curves and shaded regions represent the medians and the 10th--90th percentile ranges, respectively.}
    \label{fig:mixing_tv}
\end{figure}

Next, we evaluated the dependence of the transition kernel on its initial state. Multiple chains were run independently from the same initial state, which was taken from a persistent chain used during training. At each transition step $s$, we measured the effect of the initial state using the total variation (TV) distance between the visible distribution of the trained RBM and the sample distribution for the sample points obtained after $s$ transition steps. Let the sample distribution be denoted by $Q_s$. The TV distance is then defined as
\begin{align}
D_{\mathrm{TV}}(P_{\theta},Q_s) := \frac{1}{2}\sum_{\bm{v}} |P_{\theta}(\bm{v}) - Q_s(\bm{v})|.
\label{eq:tv_distance}
\end{align}
Because the evaluation of the TV distance requires summation over all possible visible states, it is computationally feasible only when the number of visible variables is small. We therefore used only the RBM sequences trained on four datasets: GenRBM, islands, pentagon, and iris. As the number of transition steps $s$ increases, the empirical distribution of the samples is expected to approach the target distribution, thereby reducing the TV distance. If a transition kernel depends only weakly on the initial state, the TV distance is expected to decrease rapidly. Therefore, a faster decay of the TV distance indicates weaker initial-state dependence and suggests faster mixing. However, this decay does not directly measure the mixing time because it is evaluated from a specific initial state and only in the visible-variable space of the bottommost RBM of the RBM sequence. Figure~\ref{fig:mixing_tv} depicts the TV distances for the four datasets. For all four datasets, the median TV distance obtained with the proposed transition kernel decreased more rapidly than the corresponding distances obtained with BGS and DT. The proposed kernel also attained lower TV distances within the evaluated range of transition steps. These results indicate that the proposed transition kernel reduced the dependence on the initial state more rapidly than the other kernels.

\section{Conclusion} \label{sec:conclusion}

In this study, we proposed a new transition kernel defined on an RBM sequence constructed for DT. The proposed kernel performs a round trip through the RBM sequence, moving from the bottom RBM to the topmost RBM and then returning to the bottom RBM. This construction is intended to produce nonlocal transitions within a small number of transition steps by leveraging the smaller state spaces of the higher-layer RBMs. Our numerical experiments showed that the proposed transition kernel explored a broader range of states and moved between modes more frequently than BGS and DT. It also reduced the dependence on the initial state more rapidly, as indicated by the decay of the TV distance. Furthermore, learning based on the proposed transition kernel mitigated the degradation in log-likelihood observed in SMCI-based learning. These results suggest that the proposed transition kernel facilitates nonlocal transitions within a small number of steps. This property could improve the accuracy of the sampling approximation used for parameter-gradient estimation with a limited number of transition steps and thereby enhance the stability of RBM learning.

MCMC-based sampling approximations are generally biased when the Markov chain is terminated after a finite number of transition steps. Although this bias typically decreases as the number of steps increases, exact unbiasedness generally requires samples from the stationary distribution, which cannot generally be obtained simply by running a standard MCMC chain for a finite number of steps. This gap between the theoretical gradient and its practical approximation may affect the learning process. Unbiased MCMC estimators based on coupled Markov chains have recently been developed to remove the bias~\cite{Jacob2020_UMCMC}. Combining such estimators with the proposed transition kernel is a possible direction for future work. Other important directions include a theoretical analysis of the mixing properties of the proposed kernel and an investigation of how the depth and layer dimensions of the RBM sequence affect its performance.

\bibliographystyle{ieeetr}
\bibliography{refs}

\appendix

\section{Semi-Second-Order SMCI for RBMs} 
\label{app:s2-SMCI}

This appendix presents the specific formulations of semi-second-order SMCI for the expectations $\mathbb{E}_{P_{\theta}(\bm{v}, \bm{h})}[v_i]$, $\mathbb{E}_{P_{\theta}(\bm{v}, \bm{h})}[h_j]$, and $\mathbb{E}_{P_{\theta}(\bm{v}, \bm{h})}[v_i h_j]$. These expectations are required to evaluate the parameter gradients in Eqs.~\eqref{eq:grad_b}--\eqref{eq:grad_W}. The estimators for $\mathbb{E}_{P_{\theta}(\bm{v}, \bm{h})}[v_i]$ and $\mathbb{E}_{P_{\theta}(\bm{v}, \bm{h})}[h_j]$ are formulated as
\begin{align*}
\mathcal{M}^{(v)}_{i}(\bm{v}) &:= \mathbb{E}_{P_\theta(v_i,\bm{h}\mid \bm{v}_{-i})}[v_i] = \sigmoid[\phi_{i}(\bm{v};\theta)], \\
\mathcal{M}^{(h)}_{j}(\bm{h}) &:= \mathbb{E}_{P_\theta(\bm{v},h_j\mid \bm{h}_{-j})}[h_j] = \sigmoid[\psi_{j}(\bm{h};\theta)], 
\end{align*}
respectively, where $\bm{v}_{-i}\in\{0,1\}^{n-1}$ and $\bm{h}\in\{0,1\}^{m-1}$ denote the visible variables excluding $v_i$ and the hidden variables excluding $h_j$, respectively, and
\begin{align*}
\phi_{i}(\bm{v};\theta) &:= b_i + \sum_{j\in H} \bigl[\softplus(\xi_{j,i}(\bm{v};\theta) + W_{i,j}) - \softplus(\xi_{j,i}(\bm{v};\theta))\bigr],\\
\psi_{j}(\bm{h};\theta) &:= c_j + \sum_{i\in V} \bigl[\softplus(\eta_{i,j}(\bm{h};\theta) + W_{i,j}) - \softplus(\eta_{i,j}(\bm{h};\theta))\bigr].
\end{align*}
Here, $\eta_{i,j}(\bm{h};\theta) := \eta_{i}(\bm{h};\theta) - W_{i,j} h_j$ and $\xi_{j,i}(\bm{v},\theta) := \xi_{j}(\bm{v},\theta) - W_{i,j} v_i$. Moreover, we consider two estimators for $\mathbb{E}_{P_{\theta}(\bm{v}, \bm{h})}[v_i h_j]$ given by
\begin{align*}
\mathcal{C}^{(v)}_{i,j}(\bm{v}) &:= \mathbb{E}_{P_\theta(v_i,\bm{h}\mid \bm{v}_{-i})}[v_i h_j] = \sigmoid\left(\logit\bigl\{\sigmoid[\xi_{j,i}(\bm{v};\theta)] \sigmoid[\phi_{i}(\bm{v};\theta)]\bigr\} + W_{i,j}\right), \\
\mathcal{C}^{(h)}_{i,j}(\bm{h}) &:= \mathbb{E}_{P_\theta(v_i,\bm{h}\mid \bm{v}_{-i})}[v_i h_j] = \sigmoid\left(\logit\bigl\{\sigmoid[\eta_{i,j}(\bm{h};\theta)] \sigmoid[\psi_{j}(\bm{h};\theta)]\bigr\} + W_{i,j}\right),
\end{align*}
where $\logit(x) := \ln(x/(1-x))$ denotes the logit function (i.e., the inverse function of the sigmoid function), and
\begin{align*}
\phi_{i,j}(\bm{v};\theta) &:= \phi_{i}(\bm{v};\theta) - \bigl[\softplus(\xi_{j,i}(\bm{v};\theta) + W_{i,j}) - \softplus(\xi_{j,i}(\bm{v};\theta))\bigr], \\
\psi_{j,i}(\bm{h};\theta) &:= \psi_{j}(\bm{h};\theta) - \bigl[\softplus(\eta_{i,j}(\bm{h};\theta) + W_{i,j}) - \softplus(\eta_{i,j}(\bm{h};\theta))\bigr].
\end{align*}
Given the sample points $\{(\mathbf{v}^{(\nu)}, \mathbf{h}^{(\nu)}) \mid \nu=1,2,\ldots,N_{\mathrm{s}}\}$, we approximate the expectations as follows:
\begin{align*}
\mathbb{E}_{P_{\theta}(\bm{v}, \bm{h})}[v_i] &\approx \frac{1}{N_{\mathrm{s}}}\sum_{\nu=1}^{N_{\mathrm{s}}} \mathcal{M}^{(v)}_{i}(\mathbf{v}^{(\nu)}), \\
\mathbb{E}_{P_{\theta}(\bm{v}, \bm{h})}[h_j] &\approx \frac{1}{N_{\mathrm{s}}}\sum_{\nu=1}^{N_{\mathrm{s}}} \mathcal{M}^{(h)}_{j}(\mathbf{h}^{(\nu)}), \\
\mathbb{E}_{P_{\theta}(\bm{v}, \bm{h})}[v_i h_j] &\approx \frac{1}{N_{\mathrm{s}}}\sum_{\nu=1}^{N_{\mathrm{s}}} \frac{\mathcal{C}^{(v)}_{i,j}(\mathbf{v}^{(\nu)}) + \mathcal{C}^{(h)}_{i,j}(\mathbf{h}^{(\nu)})}{2}, 
\end{align*}
respectively.

\section{Transition Probability Matrix for BGS} \label{app:trans_matrix}

In this appendix, we discuss the transition probability matrix for BGS. The relaxation time can be evaluated by estimating the second-largest eigenvalue of the transition probability matrix. We consider the transition probability matrix corresponding to the BGS kernel in Eq.~\eqref{eq:bgs}, $\bm{P}\in\mathbb{R}^{2^n\times 2^n}$, whose elements are defined by $P_{a,b} := T_{\theta}^{\mathrm{BGS}}(\mathbf{v}^{(b)}\mid\mathbf{v}^{(a)})$ for $a,b\in\{1,2,\ldots,2^n\}$. Here, $\mathcal{V}:=\{\mathbf{v}^{(1)},\mathbf{v}^{(2)},\ldots,\mathbf{v}^{(2^n)}\}$ denotes an enumeration of all visible states in $\{0,1\}^n$, and $\mathbf{v}^{(a)}\in\mathcal{V}$ is the $a$-th visible state. We define a matrix similar to $\bm{P}$ by $\bm{S}:=\bm{D}^{1/2}\bm{P}\bm{D}^{-1/2}$, where 
\begin{align*}
\bm{D}:=\mathrm{diag}\bigl(P_{\theta}(\mathbf{v}^{(1)}),P_{\theta}(\mathbf{v}^{(2)}),\ldots,P_{\theta}(\mathbf{v}^{(2^n)})\bigr).
\end{align*}
Here, $\bm{D}^{1/2}$ and $\bm{D}^{-1/2}$ are diagonal matrices whose $a$-th diagonal entries are $\sqrt{P_{\theta}(\mathbf{v}^{(a)})}$ and $1/\sqrt{P_{\theta}(\mathbf{v}^{(a)})}$, respectively. Since the probability matrix $\bm{P}$ satisfies the detailed balance condition (i.e., $\bm{D}\bm{P} = \bm{P}^\top \bm{D}$), we have
\begin{align*}
\bm{S}^\top &= \bm{D}^{-1/2}\bm{P}^\top\bm{D}^{1/2} = \bm{D}^{-1/2}\bm{D}\bm{P}\bm{D}^{-1/2} = \bm{S}.
\end{align*}
Moreover, for any real vector $\bm{x}\in\mathbb{R}^{2^n}$, we have
\begin{align*}
\bm{x}^\top \bm{S} \bm{x} 
&= \sum_{a=1}^{2^n}\sum_{b=1}^{2^n} x_a x_b \sqrt{\frac{P_\theta(\mathbf{v}^{(a)})}{P_\theta(\mathbf{v}^{(b)})}} \sum_{\bm{h}} P_\theta(\mathbf{v}^{(b)}\mid\bm{h}) P_\theta(\bm{h}\mid\mathbf{v}^{(a)}) \\
&= \sum_{a=1}^{2^n}\sum_{b=1}^{2^n} x_a x_b \sqrt{\frac{P_\theta(\mathbf{v}^{(a)})}{P_\theta(\mathbf{v}^{(b)})}} \sum_{\bm{h}} P_\theta(\mathbf{v}^{(b)}\mid\bm{h}) \frac{P_\theta(\mathbf{v}^{(a)}\mid\bm{h}) P_\theta(\bm{h})}{P_\theta(\mathbf{v}^{(a)})} \\
&= \sum_{\bm{h}} P_\theta(\bm{h}) \left(\sum_{a=1}^{2^n} x_a \frac{P_\theta(\mathbf{v}^{(a)}\mid\bm{h})}{\sqrt{P_\theta(\mathbf{v}^{(a)})}}\right)^2 \ge 0.
\end{align*}
Therefore, $\bm{S}$ is symmetric and positive semidefinite, meaning that all eigenvalues of $\bm{S}$ are nonnegative real numbers. As any probability matrix has an eigenvalue equal to $1$, the eigenvalues of $\bm{S}$ can be ordered in descending order as
\begin{align}
1=\lambda_1 \ge \lambda_2 \ge \cdots \ge \lambda_{2^n} \ge 0. \label{eq:eigenvalues}
\end{align}
As $\bm{S}$ is similar to $\bm{P}$, the eigenvalues of $\bm{P}$ are also given by Eq.~\eqref{eq:eigenvalues}. The relaxation time of BGS is defined as
\begin{align*}
t_{\mathrm{rel}}:=\frac{1}{1-\lambda_2}.
\end{align*}

We next consider the computation of the second-largest eigenvalue $\lambda_2$, which is used to evaluate the relaxation time. Here, for $\bm{u}_1:=\bm{D}^{1/2}\bm{1}_{2^n}$, we have $\bm{S}\bm{u}_1=\bm{u}_1$ and $\bm{u}_1^\top\bm{u}_1=1$. Therefore, $\bm{u}_1$ is the eigenvector of $\bm{S}$ corresponding to the largest eigenvalue $\lambda_1$. By Rayleigh--Ritz theorem, the eigenvector corresponding to the second-largest eigenvalue $\lambda_2$, $\bm{u}_2$, is obtained as
\begin{align*}
\bm{u}_2 = \argmax_{\substack{ \bm{q}\in\mathbb R^{2^n}\\ \bm{q}^\top \bm{q}=1\\ \bm{q}^\top\bm{u}_1=0 }} \bm{q}^\top\bm{S}\bm{q}.
\end{align*}
The corresponding eigenvalue is $\lambda_2 = \bm{u}_2^\top \bm{S} \bm{u}_2$. We denote the orthogonal projection matrix that removes the component in the direction of $\bm{u}_1$ as $\bm{\Pi}_{\perp} := \bm{I}_{2^n} - \bm{u}_1\bm{u}_1^{\top}$, and the orthogonality constraint for $\bm{u}_1$ can be then eliminated as
\begin{align}
\bm{u}_2 = \argmax_{\substack{ \bm{q}\in\mathbb R^{2^n} \\ \bm{q}^\top \bm{q}=1 }} \bm{q}^\top\bm{\Pi}_{\perp}\bm{S}\bm{\Pi}_{\perp}\bm{q}.
\label{eq:eigenvector}
\end{align}
Thus, $\bm{u}_2$ can be obtained as an eigenvector corresponding to the largest eigenvalue of $\bm{\Pi}_{\perp}\bm{S}\bm{\Pi}_{\perp}$.

Hereafter, we describe how $\lambda_2$ was estimated in the numerical experiment in Section~\ref{ssec:demonstrate_failure}. We first preprocess $\bm
D$ and $\bm{\Pi}_{\perp}$. We subsequently evaluate the transition probability matrix $\bm{P}$ using a sampling approximation. For each visible state $\mathbf{v}^{(a)} \in \mathcal{V}$, we independently perform $N_2=2^{15}$ one-step BGS transitions:
\begin{align*}
\mathbf{h}_{i}^{(a)} \sim P_{\theta}(\bm{h}\mid\mathbf{v}^{(a)}),\quad\mathbf{v}_{i}^{(a)} \sim P_{\theta}(\bm{v}\mid\mathbf{h}_{i}^{(a)}),
\end{align*}
for $i=1,2,\ldots,N_2$. The transition probability from $\mathbf{v}^{(a)}$ to $\mathbf{v}^{(b)}$ is then estimated by
\begin{align*}
\hat{P}_{a,b} := \frac{1}{N_2} \sum_{i=1}^{N_2} \delta(\mathbf{v}_{i}^{(a)}, \mathbf{v}^{(b)}).
\end{align*}
We then estimate the matrix $\bm{A}:=\bm{D}\bm{P}$, which is symmetric because BGS satisfies the detailed balance condition, as
\begin{align*}
\hat{\bm{A}}:=\frac{1}{2}\left(\bm{D}\hat{\bm{P}} +(\bm{D}\hat{\bm{P}})^\top\right).
\end{align*}
Since $\mathbb{E}[\hat{\bm{P}}]=\bm P$ and $\bm{A}=\bm{A}^\top$, it holds $\mathbb{E}[\hat{\bm{A}}]=\bm{A}$. Thus, $\hat{\bm{A}}$ is a symmetric unbiased estimator of $\bm{A}$. Using $\hat{\bm{A}}$, the estimator of $\bm{S}$ is obtained by $\hat{\bm{S}}:=\bm{D}^{-1/2}\hat{\bm{A}}\bm{D}^{-1/2}$, and we estimate $\bm{u}_2$ by obtaining the eigenvector corresponding to the largest eigenvalue of $\bm{\Pi}_{\perp}\hat{\bm{S}}\bm{\Pi}_{\perp}$. Using the resulting eigenvector $\hat{\bm{u}}_2$, we estimate $\lambda_2$ by $\hat{\lambda}_2:=\hat{\bm{u}}_2^\top \hat{\bm{S}}\hat{\bm{u}}_2$. Note that the resulting estimator $\hat{\lambda}_2$ is generally biased and is not guaranteed to provide either a lower or an upper bound on $\lambda_2$ because the estimator $\hat{\bm{S}}$ is used both to determine $\hat{\bm{u}}_2$ and to evaluate its Rayleigh quotient.

\section{Invariant Distribution of the Proposed Transition Kernel}
\label{app:invariant_distribution}

We prove that the proposed transition kernel $T_\Theta^{\mathrm{prop}}$ leaves $P_\Theta(\bm{V})$ invariant. We consider the product distribution from the $\ell$-th RBM to the topmost RBM defined by $\Pi_{\ell}(\bm{V}_{\ell:L}) := \prod_{i=\ell}^{L} P_{\theta_i}(\bm{v}_i)$. First, the topmost kernel $T_L$  leaves $\Pi_{L}$ invariant because $T_L$ is one step of BGS and satisfies the detailed balance condition for $\Pi_{L}$. Next, we fix $\ell\in\{0,\ldots,L-1\}$ and assume that $T_{\ell+1}$ satisfies the balance condition for $\Pi_{\ell+1}$, i.e.,
\begin{align}
\sum_{\bm{V}_{\ell+1:L}}T_{\ell+1}(\bm{V}_{\ell+1:L}'\mid\bm{V}_{\ell+1:L})\Pi_{\ell+1}(\bm{V}_{\ell+1:L})
=\Pi_{\ell+1}(\bm{V}_{\ell+1:L}').
\label{eq:induction_hypothesis}
\end{align}
We then prove that the kernel $T_\ell$ leaves $\Pi_{\ell}$ invariant by following the sampling procedure of $T_\ell$ in Eq.~\eqref{eq:ell_trans} step by step. First, sampling from $P_{\theta_\ell}(\bm{h}_\ell\mid\bm{v}_\ell)$ gives
\begin{align*}
\sum_{\bm{v}_\ell}P_{\theta_\ell}(\bm{h}_\ell\mid\bm{v}_\ell) \Pi_{\ell}(\bm{V}_{\ell:L})
=P_{\theta_\ell}^{(h)}(\bm{h}_\ell)\Pi_{\ell+1}(\bm{V}_{\ell+1:L}).
\end{align*}
Since the swap kernel $S_{\ell,\ell+1}$ leaves $P_{\theta_\ell}^{(h)}P_{\theta_{\ell+1}}^{(v)}$ invariant, applying the swap kernel gives
\begin{align*}
&\sum_{\bm{h}_\ell}\sum_{\bm{v}_{\ell+1}}S_{\ell,\ell+1}(\bm{h}_\ell^+,\bm{v}_{\ell+1}^+\mid\bm{h}_\ell,\bm{v}_{\ell+1})P_{\theta_\ell}^{(h)}(\bm{h}_\ell)P_{\theta_{\ell+1}}^{(v)}(\bm{v}_{\ell+1})\Pi_{\ell+2}(\bm{V}_{\ell+2:L}) \\
&\qquad=P_{\theta_\ell}^{(h)}(\bm{h}_\ell^+)\Pi_{\ell+1}(\bm{v}_{\ell+1}^+,\bm{V}_{\ell+2:L}).
\end{align*}
Applying $T_{\ell+1}$ and using Eq.~\eqref{eq:induction_hypothesis} yields
\begin{align*}
&\sum_{\bm{v}_{\ell+1}^+} \sum_{\bm{V}_{\ell+2:L}} T_{\ell+1}(\hat{\bm{v}}_{\ell+1},\bm{V}_{\ell+2:L}'\mid\bm{v}_{\ell+1}^+,\bm{V}_{\ell+2:L}) P_{\theta_\ell}^{(h)}(\bm{h}_\ell^+)\Pi_{\ell+1}(\bm{v}_{\ell+1}^+,\bm{V}_{\ell+2:L})\\
&\qquad=P_{\theta_\ell}^{(h)}(\bm{h}_\ell^+)\Pi_{\ell+1}(\hat{\bm{v}}_{\ell+1},\bm{V}_{\ell+2:L}').
\end{align*}
Applying the swap kernel in the downward direction gives
\begin{align*}
&\sum_{\bm{h}_\ell^+}\sum_{\hat{\bm{v}}_{\ell+1}} S_{\ell,\ell+1}(\bm{h}_\ell^-,\bm{v}_{\ell+1}'\mid\bm{h}_\ell^+,\hat{\bm{v}}_{\ell+1}) P_{\theta_\ell}^{(h)}(\bm{h}_\ell^+)\Pi_{\ell+1}(\hat{\bm{v}}_{\ell+1},\bm{V}_{\ell+2:L}') \\
&\qquad=P_{\theta_\ell}^{(h)}(\bm{h}_\ell^-)\Pi_{\ell+1}(\bm{V}_{\ell+1:L}').
\end{align*}
Finally, sampling from $P_{\theta_\ell}(\bm{v}_\ell\mid\bm{h}_\ell^-)$ gives
\begin{align*}
\sum_{\bm{h}_\ell^-}P_{\phi_\ell}(\bm{v}_\ell'\mid\bm{h}_\ell^-) P_{\theta_\ell}^{(h)}(\bm{h}_\ell^-)\Pi_{\ell+1}(\bm{V}_{\ell+1:L}') = \Pi_\ell(\bm{V}_{\ell:L}').
\end{align*}
Therefore, $T_\ell$ leaves $\Pi_\ell$ invariant. By mathematical induction, the proposed transition kernel leaves $\Pi_0(\bm{V})$ (i.e., $P_{\Theta}(\bm{V})$) invariant, as does the DT kernel.

\end{document}